\documentclass[conference]{IEEEtran}
\IEEEoverridecommandlockouts
\usepackage[square,numbers]{natbib}
\usepackage{dblfloatfix}
\usepackage{wrapfig}
\usepackage{graphicx}
\usepackage{amsmath}
\usepackage{amssymb}
\usepackage{booktabs}
\usepackage{multirow}
\usepackage{multicol}
\usepackage{adjustbox}
\usepackage{subcaption}
\usepackage{enumitem}
\usepackage{breqn}
\usepackage[breaklinks=true]{hyperref}
\newcount\Comments  % 0 suppresses notes to selves in text, 1 to show comments
\usepackage{color}
\newcommand{\kibitz}[2]{\ifnum\Comments=1\textcolor{#1}{#2}\fi}

\usepackage{amsmath,amssymb,amsfonts}
\usepackage{algorithmic}
\usepackage{graphicx}
\usepackage{textcomp}
\usepackage{xcolor}
\def\BibTeX{{\rm B\kern-.05em{\sc i\kern-.025em b}\kern-.08em
    T\kern-.1667em\lower.7ex\hbox{E}\kern-.125emX}}
\begin{document}

\title{\textsc{ColorSense}: A Study on Color Vision in Machine Visual Recognition}

\author{
\IEEEauthorblockN{Ming-Chang Chiu}
\IEEEauthorblockA{\textit{USC}\\
mingchac@usc.edu}
\and
\IEEEauthorblockN{ Yingfei Wang}
\IEEEauthorblockA{\textit{USC}\\
yingfei@usc.edu}
\and
\IEEEauthorblockN{Derrick Eui Gyu Kim}
\IEEEauthorblockA{\textit{USC}\\
euigyuki@usc.edu}
\and
\IEEEauthorblockN{Pin-Yu Chen}
\IEEEauthorblockA{\textit{IBM Research}\\
pin-yu.chen@ibm.com}
\and
\IEEEauthorblockN{Xuezhe Ma}
\IEEEauthorblockA{\textit{USC}\\
xuezhema@usc.edu}

}

\maketitle

\begin{abstract}
Color vision is essential for human visual perception, but its impact on machine perception is still underexplored. There has been an intensified demand for understanding its role in machine perception for safety-critical tasks such as assistive driving and surgery but lacking suitable datasets. To fill this gap, we curate multipurpose datasets \textsc{ColorSense}, by collecting \textbf{110,000 non-trivial human annotations} of foreground and background color labels from popular visual recognition benchmarks. To investigate the impact of color vision on machine perception, we assign each image a color discrimination level based on its dominant foreground and background colors and use it to study the impact of color vision on machine perception. We validate the use of our datasets by demonstrating that the level of color discrimination has a dominating effect on the performance of mainstream machine perception models. Specifically, we examine the perception ability of machine vision by considering key factors such as model architecture, training objective, model size, training data, and task complexity. Furthermore, to investigate how color and environmental factors affect the robustness of visual recognition in machine perception, we integrate our \textsc{ColorSense} datasets with image corruptions and perform a more comprehensive visual perception evaluation. We jointly analyze the impact of color vision and image corruption on machine perception. Our findings suggest that \emph{object recognition} tasks such as \emph{classification} and \emph{localization} are susceptible to color vision bias, especially for high-stakes cases such as vehicle classes, and advanced mitigation techniques such as data augmentation and so on only give marginal improvement. Our analyses highlight the need for new approaches toward the performance evaluation of machine perception models in real-world applications. Lastly, we present various potential applications of \textsc{ColorSense} such as studying spurious correlations.\footnote{Dataset is available at https://github.com/charismaticchiu/ColorSense}\footnote{This work has been accepted for publication in the IEEE Conference on Secure and Trustworthy Machine Learning (SaTML). The final version will be available on IEEE Xplore.}
\end{abstract}

\section{Introduction}
The development of deep neural networks (DNN) has been deeply rooted in inspiration drawn from neuroscience and cognitive science \cite{richards2019deep}.
For instance, convolutional neural networks (CNN) \cite{lecun1989backpropagation, he2016deep} have enjoyed great success in various tasks in computer vision by modeling the human visual cortex system. More recently, Transformer \cite{vaswani2017attention} 
has become a widely adopted architecture that models the human attention mechanism and allows better contextual understanding. It has since expanded its success beyond natural language processing to computer vision \cite{dosovitskiy2020image}. Some recent studies explore the behavioral aspects of DNN models for object recognition \cite{tuli2021convolutional, Geirhos2020BeyondAQ, hermann2020origins}. However, there are other important parts of the visual pathway of primates that remain underexplored, such as \textit{color vision} \cite{neurosci_book, biochem_texbook}, a neuroscience topic that studies how color affects human visual perception. With such a nomination, we curate new datasets to study machine visual recognition as a proxy task to connect the \textit{color vision} aspects of human visual perception with those of machine vision. 
%(more on color vision in \S\ref{sec:background}). %, and then extend the analysis to a broader visual perception setting. %\faye{"a broader visual perception setting?"}
\begin{figure*}[t]
     \centering
     \includegraphics[width=1\linewidth]{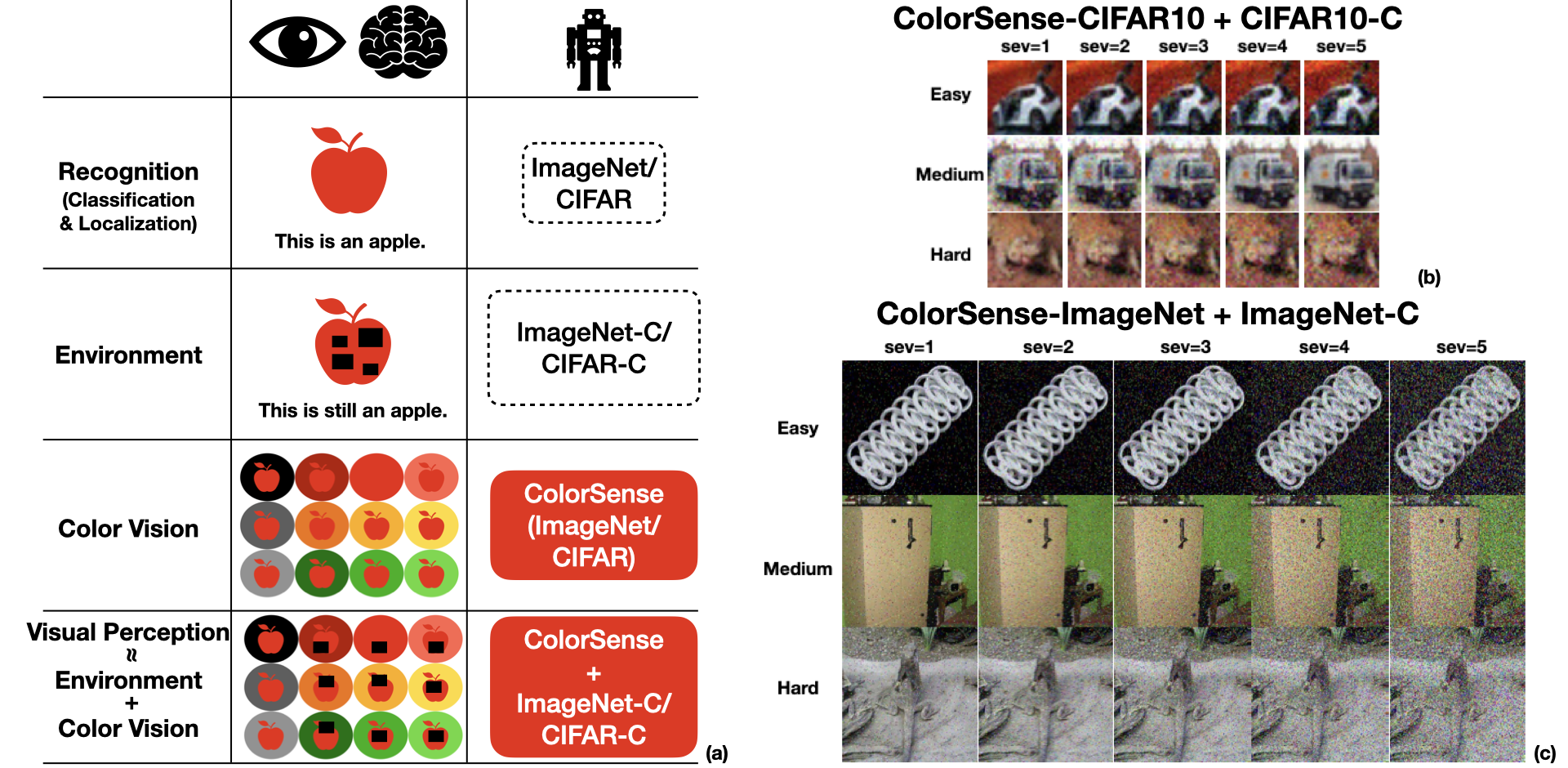}
        \caption{\textbf{Mapping of our \textsc{ColorSense} datasets to human visual aspects and examples of our \textsc{ColorSense} integrated with corrupted images (ImageNet and CIFAR10).} (a) \textsc{ColorSense} can evaluate the color vision and more complete visual perception capabilities. (b-c) Examples of \textsc{ColorSense} integrated with corrupted datasets. Each row indicates a color discrimination level.}
        \label{fig:big_picture}
\end{figure*}
% Contrast sensitivity is a fundamental property of the human visual system, which indexes the limits of what one can detect or discriminate–the window of visibility.  \cite{himmelberg2021linking}
%  The primate visual system can robustly and effortlessly discriminate among visual objects over the wide range of images that each object produces during natural vision \cite{dicarlo2012does, zoccolan2007trade, potter1976short, intraub1980presentation, rubin1992reading, logothetis1996visual, thorpe1996speed, edelman1999representation, rousselet2004parallel}
% The key computational problem of object recognition is attaining both selectivity among different visual objects and tolerance to variation in each object’s appearance. The primate visual system has solved this problem: primates robustly and effortlessly discriminate among visual objects over the wide range of images that each object produces during natural vision \cite{dicarlo2012does, zoccolan2007trade, potter1976short, intraub1980presentation, rubin1992reading, logothetis1996visual, thorpe1996speed, edelman1999representation, rousselet2004parallel}
% Retinal and LGN processing help deal with important real-world issues such as variation in luminance and contrast across each visual image \cite{kohn2007visual}
Color is a fundamental perception that we often take for granted as humans. It is only when our color vision is hindered, as in the cases of colorblindness, that we realize its true significance. Similarly, color plays a role in machine vision \cite{olah2020an, flachot2022deep, nadler2023divergences} and has a crucial impact on downstream safety-critical visual recognition tasks such as self-driving, detection of skin disease, or robotic-assisted surgery. Previous research has identified similarities between human and machine vision. For example, the last convolution layers in classical machine vision models are shown to have high color sensitivity,  similar to the most anterior occipital visual areas in primates \cite{flachot2021color}, and strong white balance can negatively impact DNNs models in image segmentation and classification tasks \cite{afifi2019else}. However, comprehensive evaluations of cutting-edge models remain limited due to the lack of well-annotated datasets that can be seamlessly integrated with modern computer vision research. To the best of our knowledge, \cite{9607588} was the only proposed dataset to date that aims to facilitate the studies on the effect of color distortion. To address this need and to enable a wider range of studies on color vision, we curate \textsc{ColorSense-CIFAR} and \textsc{ColorSence-ImageNet}. We manually separated ImageNet into eight groups (red, green, blue, gray, brown, black, white, others) based on the dominant foreground and background colors of each image. Furthermore, to study how neighboring colors help machines discriminate objects, we consult color professionals and the established standard \cite{w3c2008WCAG} to categorize the foreground/background color pairs into 3 color discrimination groups \{\textsc{Hard}, \textsc{Medium}, \textsc{Easy}\} (details in \S\ref{sec:datasets}). We call this dataset \textsc{ColorSense}, designed to quantify the \emph{sense of color} for machine vision. To achieve this goal, we evaluate models beyond the regime of using total accuracy as the sole objective and examine model performance in each stratified color discrimination groups. Similarly, on CIFAR, we also provide dominant foreground color attributes (Fig.~\ref{fig:coco_examples}(b)) to create \textsc{ColorSense-CIFAR}. We mainly focus on \textsc{ColorSense-ImageNet} in this work.

%  as evidence show that neural responses activity in visual system correlates with the increase with contrast \cite{heeger2006adaptation, kohn2007visual}, 
As a benefit, \textsc{ColorSense} can also be seamlessly integrated with out-of-distribution (OOD) datasets like ImageNet-C \cite{hendrycks2019robustness} (dubbed \textsc{ColorSense-ImageNet-C}) as a complement to broaden the analysis of color vision to include simulated environmental factors such as \textit{noisy} data shifts. For human vision, it is well-established that the primate visual system can effortlessly and robustly discriminate between various visual objects, regardless of the wide range of images that each object can produce during natural vision \cite{dicarlo2012does, zoccolan2007trade, potter1976short, intraub1980presentation, rubin1992reading, logothetis1996visual, thorpe1996speed, edelman1999representation, rousselet2004parallel}, and the robustness should extend to different surroundings or added noise \cite{manahilov2003temporal,neurosci_book,biochem_texbook}. Fig.~\ref{fig:big_picture} illustrates where \textsc{ColorSense} fits in bridging the evaluation of visual perception between humans and machines.

To gain a comprehensive understanding of color vision in machine vision, we conduct comprehensive analyses on \textsc{ColorSense} in conjunction with corrupted datasets across different model sizes and architectures, training objectives, etc., which shows the diverse capabilities of \textsc{ColorSense}. This framework provides new insights into model behaviors under different environmental and color discrimination conditions. Our study is a step toward understanding color vision in deep learning, and we believe \textsc{ColorSense} will facilitate benchmarking and evaluation efforts in different ways. In addition, we explore advanced bias mitigation and provide reasoning behind the phenomenon. Lastly, we showcase other uses of our dataset: we perform case studies on a safety-critical task to demonstrate the importance of color vision, quantify model robustness, and study fairness using color subgroups from \textsc{ColorSense}. % Lastly, we demonstrate other usages of our datasets, such as safety-critical case studies and using our color subgroups for fairness and robustness studies.

To summarize, our contributions are threefold:
\begin{itemize}[leftmargin=*]
 \item We curate \textsc{ColorSense} dataset with 110,000 human annotations to study how color vision affects computer vision models on various visual recognition tasks. % inspired by the color vision function in human vision,
 \item We conduct comprehensive analyses on factors like model size, architecture, training objectives and procedures and propose generic metrics for the color vision effect and model robustness.
 % We propose two metrics and conduct extensive analyses in three dimensions: model size, architectural choice, and task complexity.%\ming{do we want this?}
    % Both analyses provide new insights on model evaluation that extend beyond average accuracy. 
 \item We show \textsc{ColorSense}'s diverse uses, such as integrating with ImageNet-C to study the robustness of machine color vision in natural and simulated noisy environments, extended use in additional recognition tasks like object localization, and studying spurious correlations.
\end{itemize}

\section{Bridging Human Vision and Machine Vision}\label{sec:background}
 
\subsection{Human Vision Test}
 We, as humans, rely on stimuli from light that reflect on objects to our eyes to perceive objects. More importantly, our retinal cone cells respond differently to wavelength in a diverse spectrum of light \cite{neurosci_book,biochem_texbook}. For instance, three different types of cone cells respond predominantly to red, green, and blue. This \textit{color vision} mechanism helps us discriminate objects under different \textit{contrast, environments}, etc., and is an essential part of why our vision can adapt quickly \cite{neurosci_book,biochem_texbook, heeger2006adaptation} and thus become more robust \cite{dicarlo2012does, zoccolan2007trade, potter1976short, intraub1980presentation, rubin1992reading, logothetis1996visual, thorpe1996speed, edelman1999representation, rousselet2004parallel}. More concretely, the opponent-process theory, a theory that different wavelengths would oppress one type of cell, explains how psychological perception would affect our way of discriminating objects \cite{neurosci_book,biochem_texbook}. Such phenomena have important real-life applications; for example, a red apple would be more clearly identified when placed on a green background (Fig.~\ref{fig:big_picture}) than on an orange background; army soldiers would wear green and brown uniforms instead of white for better camouflage. To emphasize the distinction between visual responses to different colors \cite{abrams2007relation}, we mainly use color discrimination as our terminology in experiments. % Note that as color constancy remains the same for machines (same pixel values), color vision and color discrimination represent the same concept  in our work and will be used interchangeably. 

 To measure human visual perception ability, a common routine is to perform the \textit{Snellen's test}, where subjects need to discriminate black alphabets in a white background from a certain distance. This test measures our \textit{visual acuity}, the ability to distinguish shapes and the details of objects \cite{visual_acuity_porter, owsley1987contrast}. However, a 20/20 vision does not indicate perfect vision \cite{visual_acuity_porter, owsley1987contrast, gregory2015}, and ophthalmologists have proposed the \textit{contrast sensitivity test} and some color vision tests. The former measures the ability to discriminate finer and finer increments of light versus dark \cite{heiting2019contrast}, and the latter includes \textit{color plate test, hue test, anomaloscope test} \cite{nih2019testing} to measure the accuracy of color vision. %Experts deem contrast sensitivity test a more complete assessment of vision \cite{owsley1987contrast}. In this work, we go beyond drawing analogies to neuroscience and ophthalmology, to test the ability of machine vision in this regard.
 
 \subsection{Machine Vision Test}
 Unlike various early visual perception tests for human vision, machine vision often goes directly to recognition tasks. Although DNN models reach human-level performances on diverse visual recognition tasks \cite{BeiT, yolo7, wang2022internimage}, they are not always robust \cite{Goodfellow2015ExplainingAH,hendrycks2019robustness,chen2023book}, which means there are \textit{notable gaps to true human-aligned vision}. Hence, instead of revolving around performances on generalization errors, we take a step back and ask: \textit{What are the fundamental aspects that current studies overlooked?} Inspired by the neuroscientific literature, our answer is to study \textit{how color vision affects machine visual recognition}, as an important \textit{first step} to bridge the gap. 
 Prior works have touched on the related \textit{contrast sensitivity} concept \cite{li2022contrast, akbarinia2023contrast} and \citet{olah2020an} visualizes early CNN layers to hint color's role. But due to the lack of proper datasets, they \textit{did not} evaluate DNNs' \textit{color vision} behavior that affects visual recognition. %Our work provides datasets that help in this regard.

\section{Benchmarking for Color Vision}\label{sec:datasets}
In the section, we introduce our \textsc{ColorSense-ImageNet}, a dataset that are based on the validation sets of ImageNet \cite{deng2009imagenet}. On ImageNet validation set, we label 50,000 images' dominant colors in both the foreground and background. To enable color vision study in different setups, we also provide 10,000 color labels on CIFAR10 \cite{Krizhevsky2009LearningML}, forming \textsc{ColorSense-CIFAR10}.
With our labeling efforts, we can create color discrimination groups for measuring color discrimination ability in computer vision and visual recognition. In the following sections, we mainly focus on the evaluations of \textsc{ColorSense-ImageNet} and leave CIFAR10 results in the Appendix.

\subsection{Labeling Process}\label{sec:labeling}

\begin{figure*}[t]
     \centering
     \includegraphics[width=\textwidth]{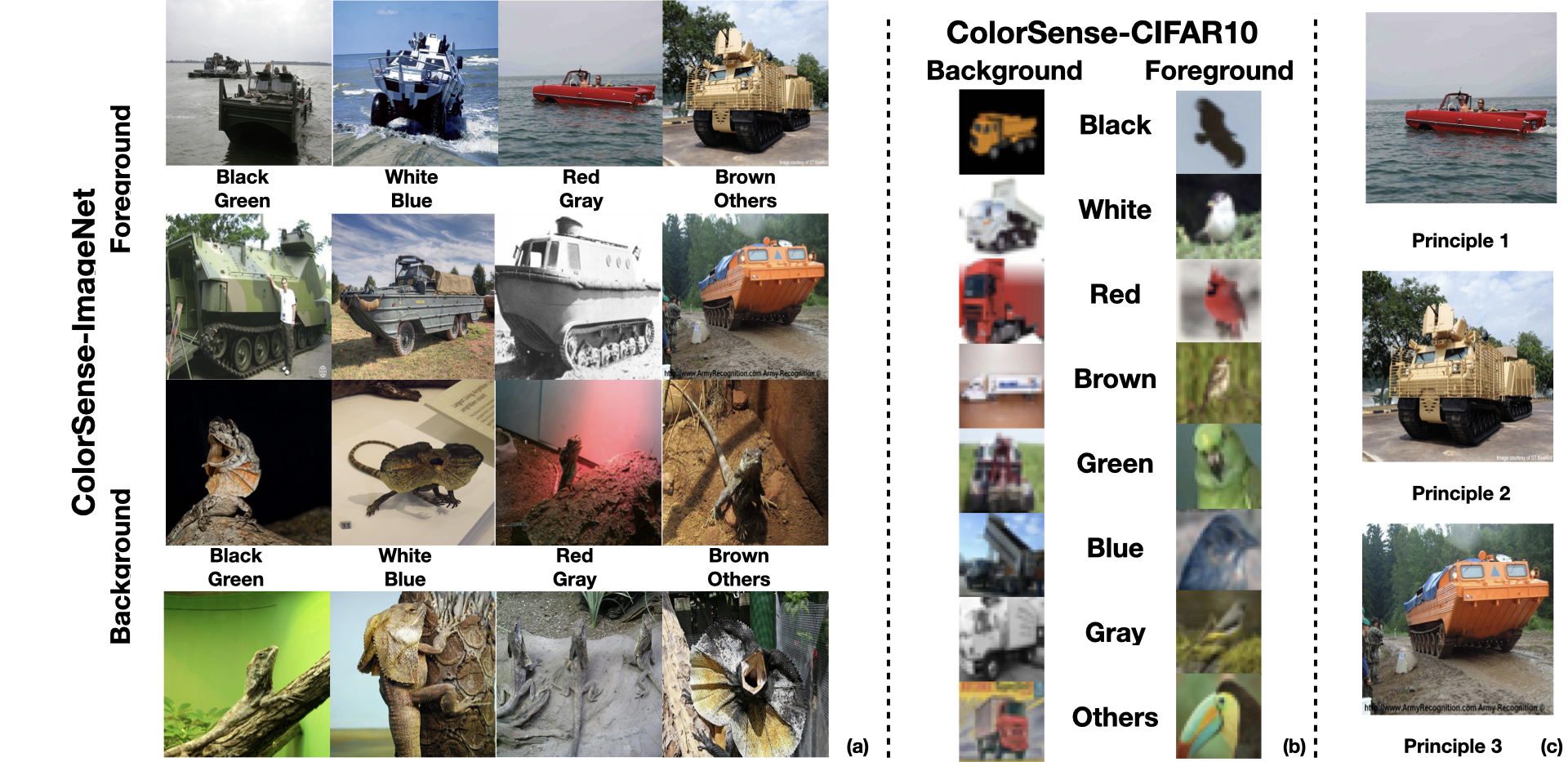}
     \caption{\textbf{Examples of our \textsc{ColorSense} datasets and labeling principles.} (a) Top two rows are \textsc{ColorSense-ImageNet-Foreground}, and the bottom two rows are \textsc{ColorSense-ImageNet-Background}. In conjunction, we can define color discrimination groups (\S~\ref{sec:datasets}). (b) \textsc{ColorSense-CIFAR10}. (c) Examples of the three labeling principles. See \S~\ref{sec:labeling} for full labeling details.}
     \label{fig:coco_examples}
\end{figure*}

\begin{enumerate}[leftmargin=*]
    % \vspace{-1mm}
    \item We label the perceived color that has the most coverage on an object. For example, in Fig.~\ref{fig:coco_examples} (c-\textit{top}), the car is almost red everywhere so we label foreground as ``red.''
    % \vspace{-2mm}
    \item When two or three colors have significant coverage, we choose the color that has more coverage. We label the example in Fig.~\ref{fig:coco_examples} (c-\textit{mid}) as ``brown'', though the black track is also significant.
    % \vspace{-2mm}
    \item When the perceived color does not belong to our categories, or when multiple colors appear on the object and none is larger than the rest, we put it in the ``others'' category (Fig.~\ref{fig:coco_examples} (c-\textit{bottom})).
\end{enumerate}

A similar procedure is applied to label the dominant background color.

% \begin{figure}[t]
%      \centering
%      \includegraphics[width=0.8\linewidth]{figures/CoCo-figures.002.png}
%         \caption{\textbf{Examples of our \textsc{ColorSense} datasets and labeling principles.} (a) Top two rows are \textsc{ColorSense-ImageNet-Foreground} and the bottom two rows are \textsc{ColorSense-ImageNet-Background}. In conjunction, we can define color discrimination groups (\S~\ref{sec:datasets}). (b) \textsc{ColorSense-CIFAR10}. (c) Examples of the three labeling principles. See \S~\ref{sec:labeling} for full labeling details.}
%         \label{fig:coco_examples}
% \end{figure}

\subsection{Defining Color Discrimination Groups}\label{sec:cd_group}
We follow the Web Content Accessibility Guidelines (WCAG) \cite{w3c2008WCAG} and consult experts from the School of Fine Arts to assign all possible color pairs into three color discrimination (CD) groups, \{\textsc{Hard}, \textsc{Medium}, \textsc{Esay}\}. The \textsc{Hard} groups consist of images with the same foreground and background colors. The \textsc{Easy} group includes color pairs with a score larger than the WCAG Minimum Contrast Success Criterion 1.4.3 \cite{w3c2008WCAG} and they also correspond to the top one-third scores ranked among all pairs; the rest goes into \textsc{Medium}. Tab.~\ref{tab:contrast_group_stats} (in Appendix) presents the statistics for the number of color pairs in each CD group and the number of test images in each group for \textsc{ColorSense} datasets. We discuss the label quality and provide details of the color pairs for each color discrimination group in the Appendix. % Also, for better visualizations, we omit the ``others" bars, which do not affect the analyses, nor are they our foci.  

\section{Evaluation: Image Classification}

\subsection{Fundamental Questions}\label{sec:questions}
\paragraph{Does color vision affect DNN models as it affects humans?} While DNNs are artificial models of neurons in the brain, some of them are designed to mimic the structure and function of the visual cortex \cite{lecun1989backpropagation}. As such, it is possible that these models possess some form of color discrimination, similar to how humans perceive color. This hypothesis raises the question of whether color vision affects DNN models and how this could impact their performance in various applications. Further research is needed to explore the relationship between color vision and DNN models and to uncover any potential benefits or limitations. Prior works show most humans are robust to color vision groups, but the decisions vary in response time (we conduct a small-scale human study to re-verify it in the Appendix) \cite{dicarlo2012does, zoccolan2007trade, potter1976short, intraub1980presentation, rubin1992reading, logothetis1996visual, thorpe1996speed, edelman1999representation, rousselet2004parallel,Geirhos2020BeyondAQ}, whereas for machines, the distinction could be reflected in the form of performance difference. 

\paragraph{Does architecture matter?} The choice of deep learning architecture can have a significant impact on the performance and efficiency of a model \cite{liu2022convnet,he2016deep,liu2021swin,liu2022swin,sandler2018mobilenetv2}. CNNs, such as ResNets, have been the go-to architecture for computer vision tasks, thanks to their ability to learn spatial features from images. However, the emergence of newer architectures, such as the ViTs, has challenged this dominance. Unlike traditional CNNs, ViT uses self-attention mechanisms to capture global image features, allowing it to outperform CNNs on some tasks and claim less inductive bias. The debate over whether architecture matters and which one to choose, continues to be an active area of research and development. We compare diverse architectures to answer this question and hope to shed light on this area in terms of color vision.

\paragraph{Does model size matter?}  The relationship between model size and robustness is an intriguing topic of research in the field of deep learning. \citet{augmix} showed that larger models tend to be more robust to added noises. Furthermore, human vision is known to be robust to color variations \cite{dicarlo2012does, zoccolan2007trade, potter1976short, intraub1980presentation, rubin1992reading, logothetis1996visual, thorpe1996speed, edelman1999representation, rousselet2004parallel,Geirhos2020BeyondAQ}, which raises the question of whether larger DNNs are also more robust to color variations. Exploring the relationship between model size, robustness, and color variation could lead to useful insights into the functioning of DNNs and their potential applications. We compare (1) architectures of the same family and (2) models of similar sizes but different architectures to answer this question in our problem setup.

\begin{figure*}[htbp]
         \centering
         \includegraphics[width=\textwidth]{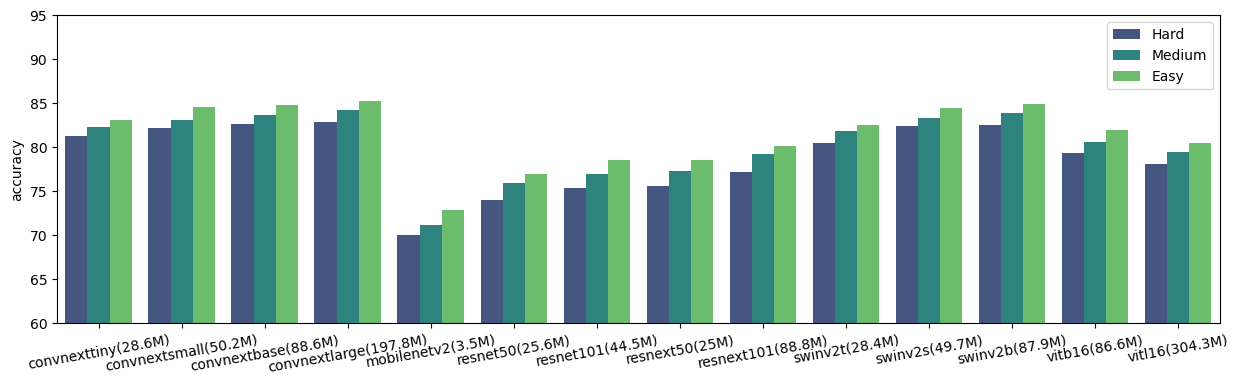} %\vspace{-3mm}
     \caption{\textbf{Testing models with different architectures and model sizes (ImageNet).} We observe similar trends in the three CD groups (Hard, Medium, Easy) across all models, showing the effect of color vision is universal across architectures and model sizes.}
     \label{fig:bar_imagenet}
     % \vspace{-5mm}
\end{figure*}
% \begin{figure*}[t]
%      \begin{subfigure}[b]{0.48\textwidth}
%          \centering
%          \includegraphics[width=\textwidth]{figures/fig41_imagenet_natrual_modelSize_barplot_rgbDict0507_threshold4_25M.pdf}
%         \caption{Models with 25M parameters}
%      \end{subfigure}%\hfill
%     %  \hspace{.9em}
%      \begin{subfigure}[b]{0.48\textwidth}
%          \centering
%          \includegraphics[width=\textwidth]{figures/fig41_imagenet_natrual_modelSize_barplot_rgbDict0507_threshold4_88M.pdf}
%         \caption{Models with 88M parameters}
%      \end{subfigure}%\hfill
%      \caption{\textbf{Testing models with different architectures and model sizes (ImageNet).} We observe similar trends across the three difficulty CD groups, showing the effect of color vision is universal across architecture and model sizes.}
%      \label{fig:bar_imagenet}
%      \vspace{-5mm}
% \end{figure*}
% \paragraph{Would pre-training change anything?} Pre-training can expose models to millions of images prior to training, and thus we presume they will be more robust to color contrast. In this work we use weights pre-trained on ImageNet.

\subsection{Main Findings}\label{sec:results_imagenet}
To answer the three fundamental questions in Sec.~\ref{sec:questions}, we conduct experiments with a diverse set of models trained on ImageNet (model weights are fixed from PyTorch). To account for the wide range of application setups, we select both classic and modern CNNs such as Resnet \cite{he2016deep}, Resnext \cite{xie2017aggregated}, Convnext \cite{liu2022convnet} and MobileNetV2 \cite{sandler2018mobilenetv2}, and vision transformers such as ViT \cite{dosovitskiy2020image} and Swin Transformers \cite{liu2021swin,liu2022swin}.

\paragraph{DNNs are deeply affected by color vision} 
We observe a consistent performance increase from the \textsc{Hard} CD group to the \textsc{Easy} group (Fig.~\ref{fig:bar_imagenet}) across all the aforementioned architectures. We perform \textit{Paired t-Tests} between the CD groups across all models. We pool the \textsc{Hard} group accuracies from all models and \textsc{Easy} group accuracies from all models and then performed the statistical tests on these two sets of data. The p-value for \textsc{Hard-Medium} is $1.39e$-$6$, \textsc{Hard-Easy} is $8.34e$-$7$, and \textsc{Medium-Easy} is $9.87e$-$6$, which are all statistically significant. We also perform a \textit{Page’s Trend Test} for increasing means ($H_0: \mu_{Hard} = \mu_{Medium} = \mu_{easy}, H_A: \mu_{Hard} < \mu_{Medium} < \mu_{Easy}$) and the p-value is $5.95e$-$7$, which is also statistically significant. These results suggest that the DNNs do have color vision ability-related issues, and color discrimination abilities should be taken into account in the future when designing and evaluating computer vision models. Second, it suggests that there is room to improve the performance of DNNs on color-based tasks by developing models that can better discriminate between colors. These implications point out a deficiency of DNNs models and shed light on an area of focus for the improvement of next-generation computer vision applications.

\paragraph{Model size and architecture do not add obvious robustness to color vision} 
For each model architecture, we evaluate its multiple variations based on size. To our surprise, Fig.~\ref{fig:bar_imagenet} shows that the performance gaps between the \textsc{Easy} and \textsc{Hard} groups do not obviously decrease as model size increases, suggesting the impact of the color discrimination group remains almost unchanged. In addition, we compare model architectures while controlling for model size; for example, models with 88M parameters (\texttt{Convnext_base, Resnext101, SwinV2_b, ViT16_b}) show similar performance gaps. While model size and architecture are essential factors to consider when designing and evaluating DNNs for various computer vision tasks, they may matter less when it comes to color vision discrimination --- we observe an upward trend in performance towards the \textsc{Easy} group regardless of the architecture or size of the model. These suggest that human-aligned color discrimination ability may not be inherent in DNNs, and developing specialized models or techniques to address color vision discrimination may be necessary.
% Moreover, to our surprise, \texttt{ViT_l} is worse than its much smaller version.

\subsection{Advanced Analyses and Ablations}\label{sec:ablations}
We have identified the effect of color vision in DNNs. To comprehensively understand its relationship with DNNs and potential mitigation, we consider additional factors across the entire deep learning pipeline, encompassing advanced training methods and training data, and evaluate foundation models like CLIPs. Further, we explore the color vision effect in real-world scenarios such as high-stake applications and environmental factors. We also propose metrics to evaluate model robustness to color vision effect. Finally, to identify potential causes of the phenomenon, we perform controlled experiments on luma and chroma, and conduct frequency analysis. 

\begin{figure}[htbp]
    \centering
    \includegraphics[width=0.45\textwidth]{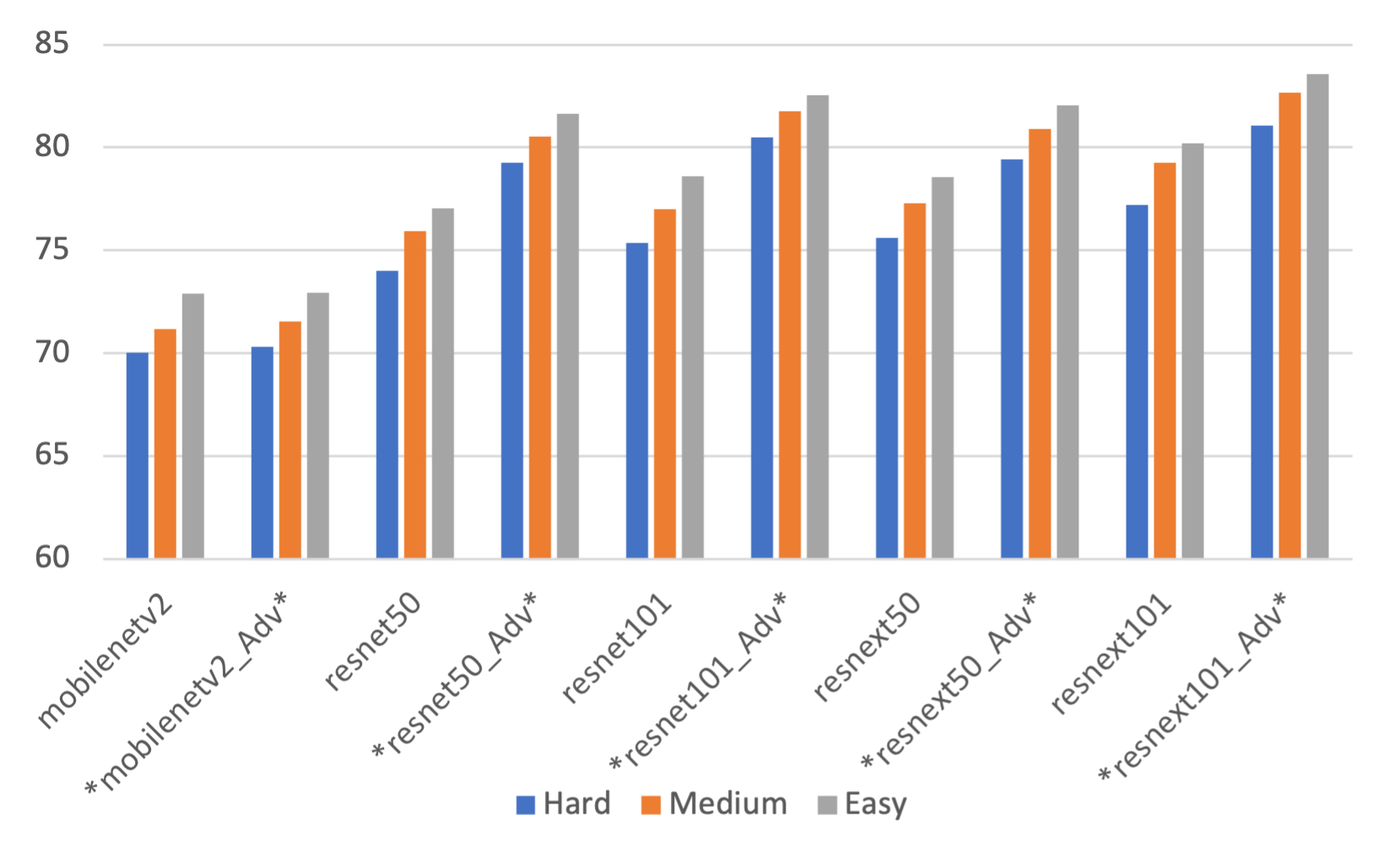}
    \caption{\textbf{Color vision performance of standard and advanced procedure-trained models.} Advanced training only makes models slightly more robust to color discrimination (also see Tab.~\ref{tab:AG}).}
    \label{fig:normal_vs_adv}
\end{figure}

\paragraph{Advanced training setups improve color vision bias marginally}
Recent works have shown that specific training recipes, such as a mix of long training epochs and complex data augmentations, can enhance model robustness and mitigate model biases \cite{chiu2022better,Vryniotis2021,wightman2021resnet,Cubuk2018AutoAugmentLA,Yun2019CutMixRS,openood}. However, our results suggest no significant changes in overall robustness to color discrimination groups (Fig.~\ref{fig:normal_vs_adv}), as measured by \textit{absolute gap} (AG), $AG = |acc_{max} - acc_{min}|$, where $acc_{max/min}$ is the maximum/minimum accuracy of the three groups (Tab.~\ref{tab:AG}). Advanced data augmentation and training only reduce AG by an average of 0.58\% in accuracy. This bolsters our belief that the color vision phenomenon exists in machine vision.

\begin{table*}[htbp]
\caption{Average absolute gap in accuracy (\%) by architecture type and dataset.}
\begin{center}
\begin{adjustbox}{width=0.9\linewidth}
\begin{tabular}{|c|c|c|c|c|c|c|}
\hline
\textbf{Absolute Gap} & \textbf{MobileNet} & \textbf{MobileNet\_adv} & \textbf{Resnet} & \textbf{Resnet\_adv} & \textbf{Resnext} & \textbf{Resnext\_adv} \\
\hline
\textsc{ColorSense} & 2.86 & 2.62 & 3.12 & 2.21 & 2.97 & 2.56 \\
\hline
Land vehicle-only & 4.35 & 4.87 & 6.59 & 4.27 & 4.63 & 5.87 \\
\hline
\textsc{ColorSense-C} & 3.37 & n/a & 4.40 & n/a & 4.46 & n/a \\
\hline
\textbf{Absolute Gap} & \multicolumn{2}{|c|}{\textbf{Convnext}} & \multicolumn{2}{c|}{\textbf{SwinV2}} & \multicolumn{2}{c|}{\textbf{ViT}} \\
\hline
\textsc{ColorSense} & \multicolumn{2}{|c|}{2.17} & \multicolumn{2}{c|}{2.18} & \multicolumn{2}{c|}{2.48} \\
\hline
Land vehicle-only & \multicolumn{2}{|c|}{3.40} & \multicolumn{2}{c|}{4.31} & \multicolumn{2}{c|}{6.93} \\
\hline
\textsc{ColorSense-C} & \multicolumn{2}{|c|}{4.59} & \multicolumn{2}{c|}{4.40} & \multicolumn{2}{c|}{4.64} \\
\hline
\end{tabular}
\end{adjustbox}
\label{tab:AG}
\end{center}
\end{table*}

% \input{tab_gap_imagenet}
% \input{fig_table_summarize}
% \input{fig_reg_cifar100}
% \input{fig_reg_cifar10-c}

\begin{figure*}[htbp]
    \begin{subfigure}[b]{0.5\textwidth}
         \centering
         \includegraphics[width=\textwidth]{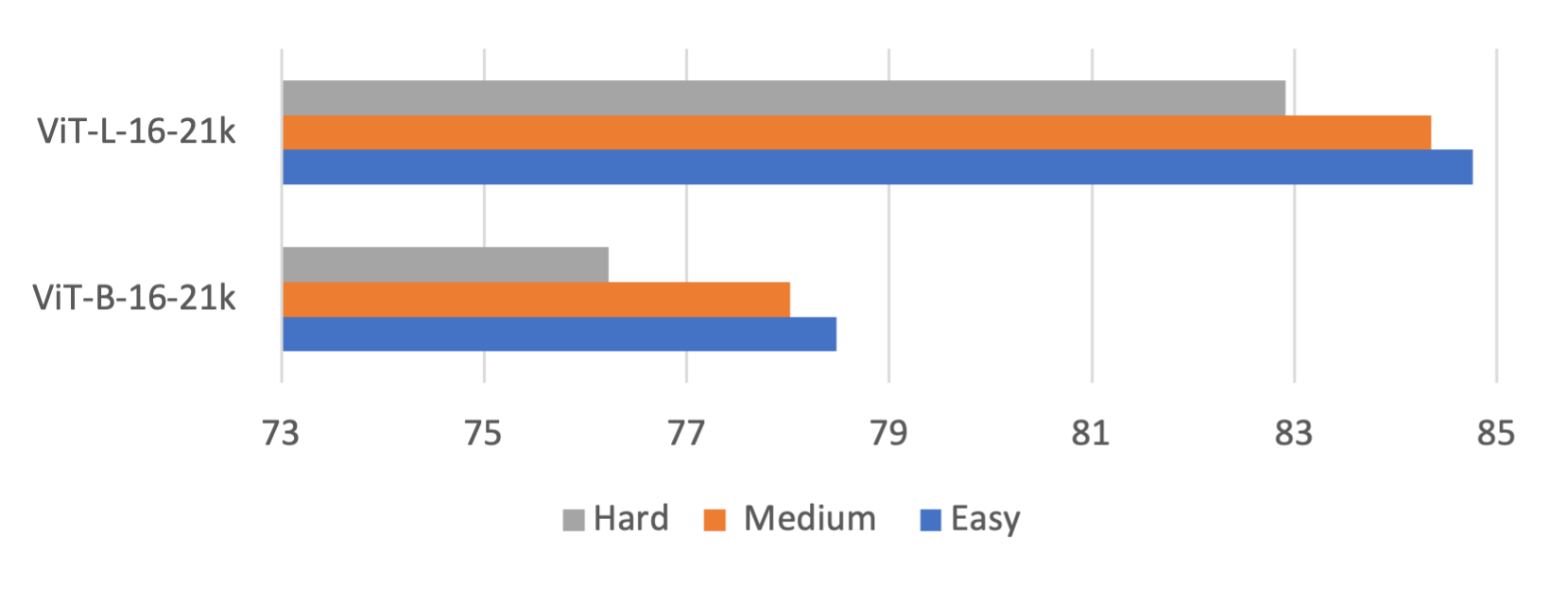}
     \end{subfigure}\hfill
     \begin{subfigure}[b]{0.5\textwidth}
         \centering
         \includegraphics[width=\textwidth]{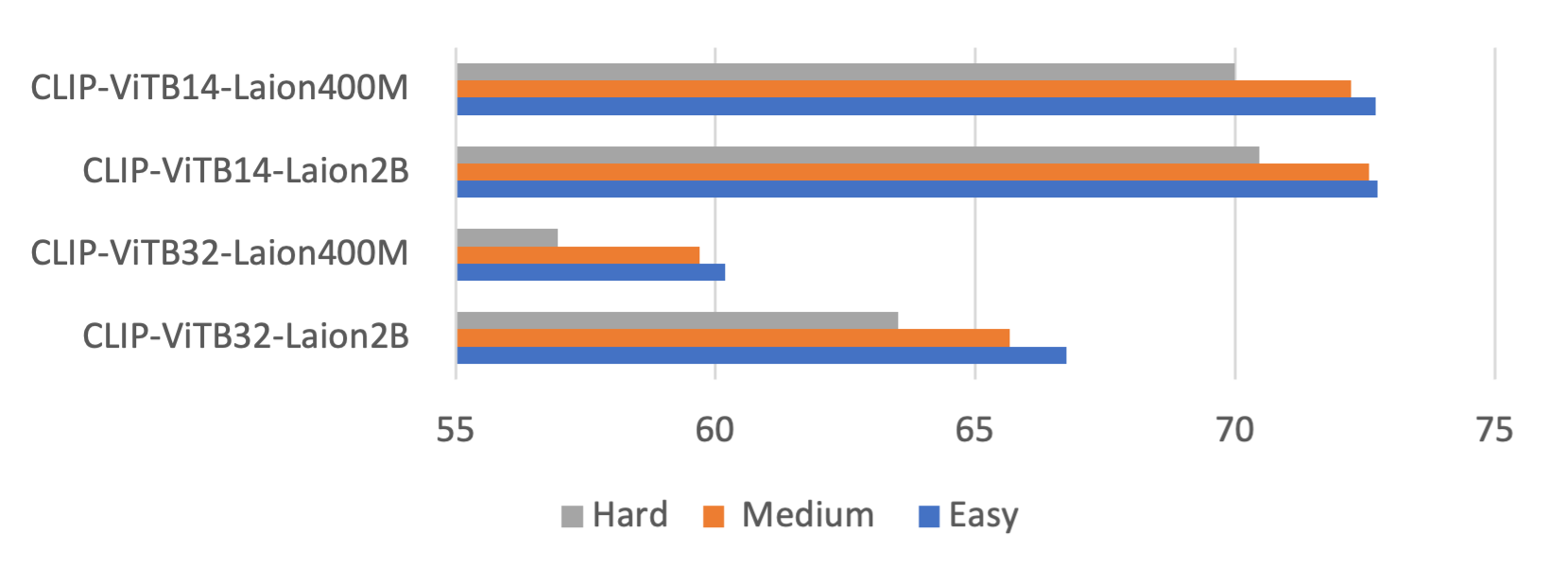}
     \end{subfigure}\vspace{-0mm}
    \caption{\textbf{Color vision in effect no matter how models are pre-trained.} \textit{Left:} ViT accuracies (\%) pre-trained on ImageNet-21k. \textit{Right:} CLIP 0-shot performances (\%) with different training data size.}
    \label{fig:compare_training_data}
    \vspace{-0mm}
\end{figure*}

% \vspace{-4mm}
\paragraph{Changing training data does not alter the color discrimination phenomenon}
To see if color discrimination can be observed when using alternative training data, we include the results of ViTs pretrained with ImageNet21k. We observe that on ViTs, the color vision effect is present (Fig.~\ref{fig:compare_training_data}-\textit{left}). 

\paragraph{Consistent patterns from self-supervised models}  We evaluate on self-supervised models like the DINO with four variants different in sizes (ViT-S/16, ViT-S/8, ViT-B/16, ViT-B/8, in this order) and report their accuracies on ImageNet in Tab.~\ref{tab:dino}. We observe a consistent pattern as in previous sections in these self-supervised models.

\begin{table}[htbp]

\caption{DINO evaluations on ImageNet}
\begin{center}
\begin{tabular}{|l|c|c|c|c|}
\hline
\textbf{CD Group} & \textbf{ViT-S/16} & \textbf{ViT-S/8} & \textbf{ViT-B/16} & \textbf{ViT-B/8} \\
\hline
Easy   & 77.90 & 80.58 & 78.78 & 80.90 \\
\hline
Medium & 76.45 & 79.25 & 77.71 & 79.41 \\
\hline
Hard   & 75.11 & 77.79 & 76.40 & 78.60 \\
\hline
\end{tabular}
\label{tab:dino}
\end{center}
\end{table}

%  \vspace{-4mm}
\paragraph{Zero-shot Foundation Models behave similarly again} We include CLIP zero-shot classification results (Fig.~\ref{fig:compare_training_data}-\textit{right}) for ViT-B32 and ViTB14 with two different training data: Laion400M and Laion2B. We observe a general trend: color vision is still in effect.
Together with advanced training and changing training data results, the three analyses suggest that \textit{training procedures do not alter the phenomenon}.

\begin{figure}[htbp]
    % \begin{minipage}[b]{0.52\textwidth}
        \centering
         \includegraphics[width=\linewidth]{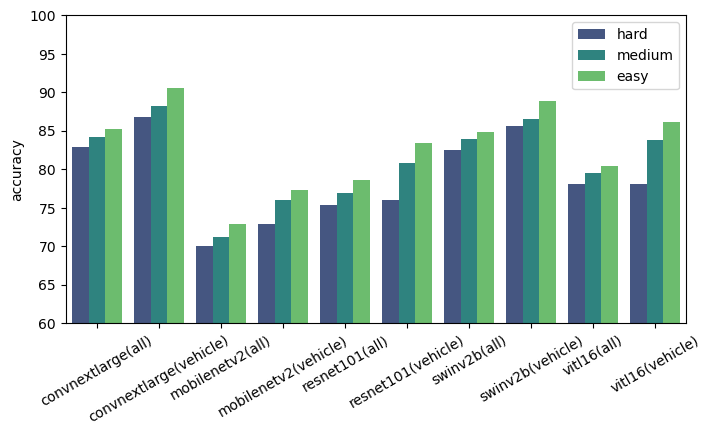}\vspace{-0mm}
    
    \captionof{figure}{\textbf{Comparisons of all vs. land vehicle classes on ImageNet.} Overall larger gaps between \textsc{Hard} and \textsc{Easy} group on vehicle classes suggest potential safety concerns when deploying such models on the road.}
    \label{fig:driving}
    % \vspace{-4.5mm}
    % \end{minipage}\hfill
    % \begin{minipage}[b]{0.45\textwidth}
    \end{figure}

% \vspace{-4mm}
\paragraph{High-stakes examples have larger gaps} Color vision is important for humans to perform high-stakes tasks such as driving or cardiovascular surgery. Prior work has shown potential lighting changes can lead to dangerous erroneous behavior \cite{pei2017deepxplore}. With the popularity of self-driving cars and the previously established knowledge about color vision in machine vision, we apply our analysis specifically to the vehicle classes in ImageNet. Ideally, we want models to be the least affected by color vision on the car-related subset.
\textit{To our surprise}, we observe models generally have \textit{even larger gaps} (p-val = $4.3e$-$2$) with average AG difference = 2.72\% (Tab.~\ref{tab:AG} \& Fig.~\ref{fig:driving}) between CD groups. It suggests vehicle classes are affected by color vision even more, and surprisingly, some larger models, such as \texttt{ViT_L} is affected more than \texttt{ViT_b}. Advanced training (Tab.~\ref{tab:AG}) only reduces the color vision gap for \texttt{Resnets} but increases for \texttt{Resnexts} and \texttt{MobileNetV2}. %Therefore, our result suggests that models like \texttt{MobileNetV2} may not be suitable to be deployed for self-driving systems. 
\textit{Our evaluations call for further color vision studies for high-stakes tasks.}

% \begin{figure*}[t]
%      \begin{subfigure}[b]{0.24\textwidth}
%          \centering
%          \includegraphics[width=\textwidth]{figures/fig44_polarplot_Allcorruption_convnextsmall_V1.png}
%         % \caption{}
%      \end{subfigure}%\hfill
%      \begin{subfigure}[b]{0.24\textwidth}
%          \centering
%          \includegraphics[width=\textwidth]{figures/fig44_polarplot_Allcorruption_swinb_V1.png}
%         % \caption{}
%      \end{subfigure}%\hfill
%      \begin{subfigure}[b]{0.24\textwidth}
%          \centering
%          \includegraphics[width=\textwidth]{figures/fig44_polarplot_Allcorruption_swinv2s_V1.png}
%         % \caption{}
%      \end{subfigure}%\hfill
%      \begin{subfigure}[b]{0.24\textwidth}
%          \centering
%          \includegraphics[width=\textwidth]{figures/fig44_polarplot_AllModels_gaussian_noise.png}
%         % \caption{}
%      \end{subfigure}%\hfill
%      \vspace{-3mm}
%      \caption{\textbf{Color discrimination capabilities with added noises by models (\textit{left three}) and by Gaussian noise (\textit{last}).}  Models of the same total performances can have very different color vision behaviors in the presence of different corruptions. \textit{Last}: CNNs see less obvious gaps between \textsc{Medium} and \textsc{Easy} group than transformers when presented Gaussian noise. These phenomena show that each architecture is built differently and is suitable for different scenarios.\vspace{-3mm}}
%      \label{fig:radar_imagenet-c}
%      \vspace{-2mm}
% \end{figure*}

\begin{figure*}[htbp]
    \centering
    \begin{subfigure}[b]{0.48\textwidth}
        \centering
        \includegraphics[width=\textwidth]{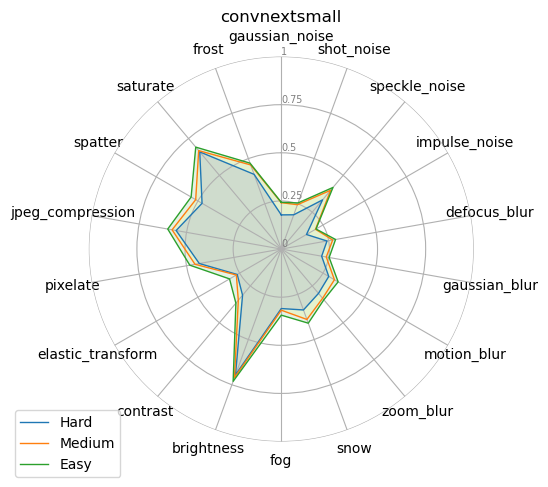}
        % \caption{}
    \end{subfigure}\hfill
    \begin{subfigure}[b]{0.48\textwidth}
        \centering
        \includegraphics[width=\textwidth]{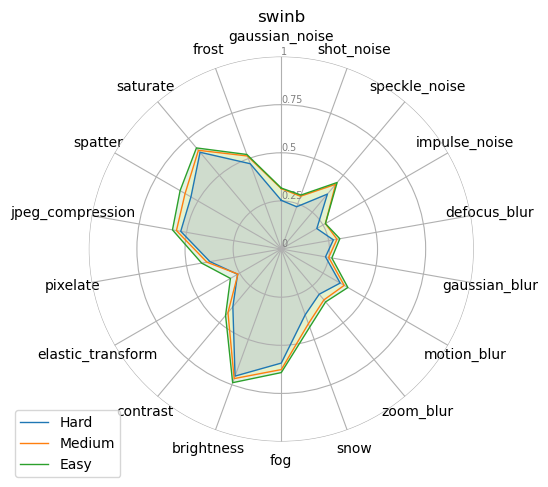}
        % \caption{}
    \end{subfigure}
    
    \vspace{0mm} % Space between the rows

    \begin{subfigure}[b]{0.48\textwidth}
        \centering
        \includegraphics[width=\textwidth]{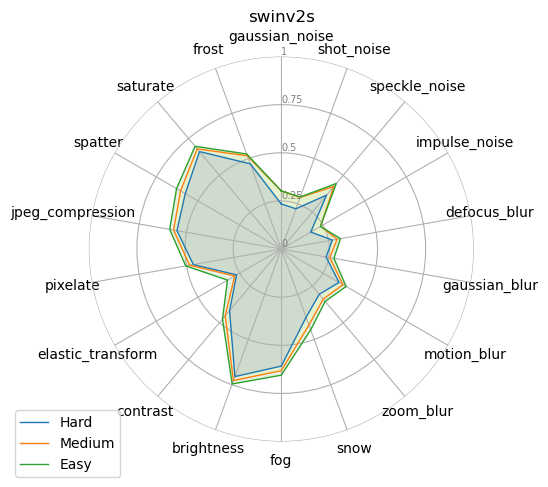}
        % \caption{}
    \end{subfigure}\hfill
    \begin{subfigure}[b]{0.48\textwidth}
        \centering
        \includegraphics[width=\textwidth]{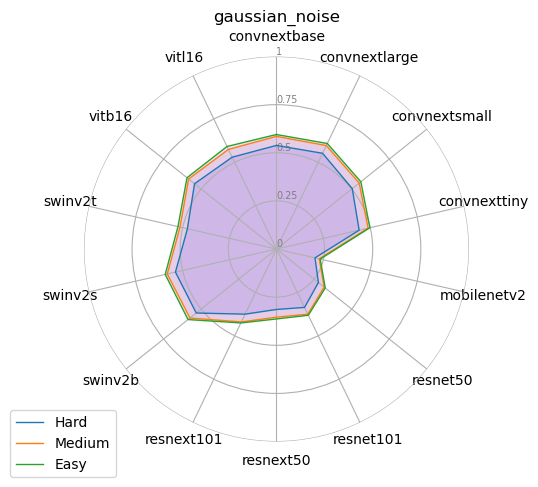}
        % \caption{}
    \end{subfigure}
    
    \vspace{-0mm}
    \caption{\textbf{Color discrimination capabilities with added noises by models and by Gaussian noise (\textit{bottom-right}).} Models of the same total performances can have very different color vision behaviors in the presence of different corruptions. \textit{Bottom-right}: CNNs see less obvious gaps between \textsc{Medium} and \textsc{Easy} group than transformers when presented Gaussian noise. These phenomena show that each architecture is built differently and is suitable for different scenarios.\vspace{-0mm}}
    \label{fig:radar_imagenet-c}
    \vspace{-0mm}
\end{figure*}

% \vspace{-4mm}
\paragraph{Environmental factors play a role}
Real-world applications of DNNs often also involve environmental factors. A benefit of \textsc{ColorSense} is the easy incorporation with OOD datasets like ImageNet-C to task models with added environmental factors and noises. We observe a consistent trend: color vision plays an important role in all types of corruption.
In Fig.~\ref{fig:radar_imagenet-c}, we show three models, \texttt{Convnext_s, Swin_b, SwinV2_s} of similar accuracies (\%) on ImageNet, 83.61, 83.58, and 83.71, respectively (less than 0.15 difference) for comparisons. Even though the generalization performances are almost the same, for each corruption type, the color vision performance is different. For instance, \texttt{Swin} \& \texttt{SwinV2}'s color discrimination capabilities are better than \texttt{Convnext_s} for all corruptions but \textit{jpeg compression}, and \texttt{SwinV2_s} is slightly better than \texttt{Swin_b} on \textit{digital} corruption such as \textit{pixelate} but worse on \textit{noise} corruptions. These findings have implications and insights for decisions such as \textit{which model we should choose?} If we expect noisy environments for some custom application, we should choose \texttt{SwinV2_s} model among the three. 

Also, we conduct case studies on specific corruptions, such as Gaussian noise (Fig.~\ref{fig:radar_imagenet-c}-\textit{bottom-right}) due to its commonality in image processing. We observe that CNNs generally have less obvious gaps in \textsc{Medium} and \textsc{Easy}  than ViTs. %It is an interesting finding for architecture choice if we expect a Gaussian noise environment.  
This analysis reveals the unique advantage of CNNs over ViTs.
Similar analyses can be done for other corruptions.
One can easily use our analysis framework to 
assess different models in more realistic settings, considering the effect of color vision.

% \begin{figure}[t]    
%         \centering
%          \includegraphics[width=0.9\textwidth]{figures/fig44_scatterplot_metrics__avg_acc_scaleby_w_std_contrast__avg_acc_scaleby_w_std_severity.png}\vspace{-2mm}
%         \captionof{figure}{\textbf{Proposed sCR \& sCVR metrics.} Overall quantified color vision robustness to color vision effect. Symbol sizes are proportional to model sizes.}
%         \label{fig:metric}
%     % \end{minipage}
%     \vspace{-3mm}
% \end{figure}
\begin{figure}[htbp]
    \centerline{
        \includegraphics[width=\columnwidth]{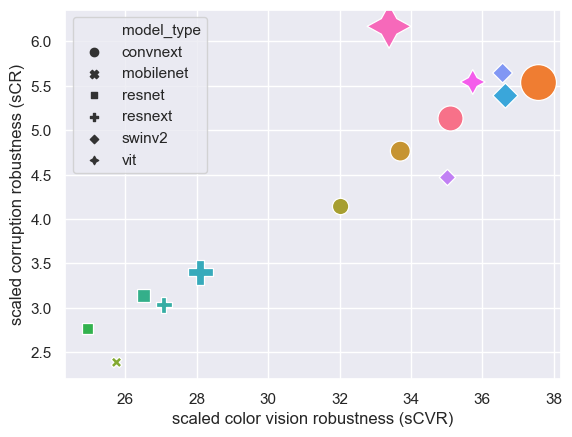}
    }
    \caption{\textbf{Proposed sCR \& sCVR metrics.} Overall quantified color vision robustness to color vision effect. Symbol sizes are proportional to model sizes.}
    \label{fig:metric}
    % \vspace{-3mm}
\end{figure}

% \vspace{-4mm}
\paragraph{Quantifying overall model robustness to color vision effect} 
To jointly consider the color vision effect and performance on corrupted dataset, we propose two metrics: \textit{scaled color vision robustness} (sCVR($\uparrow$)), $\frac{acc_{tot}}{\sigma_w}$, where $acc_{tot}$ is the total accuracy on ImageNet-C and $\sigma_w = \sqrt{\frac{\sum_{i=Easy}^{Hard} w_i (acc_i-\Bar{acc})^2}{\sum_{i=Easy}^{Hard} w_i}},$ is the weighted standard deviation across CD groups. Similarly, we define \textit{scaled corruption robustness} (sCR($\uparrow$)), $\frac{acc_{tot}}{\sigma_s}$, where $\sigma_s = \sqrt{\frac{\sum_{s=1}^{5} (acc_s-\bar{acc})^2}{\sum_{s=1}^{5} w_s}}$ (across five severity levels). Fig.~\ref{fig:metric} shows the sCVR and sCR of each model. Per our metrics, \texttt{Convnext_l} is the overall best. We also tried the $\frac{acc_{tot}}{AG}$ as a metric, and the results were similar. We observe that most models have better sCVR as their size grows. However, for ViTs, as also observed in the land vehicle classes, there is an inverse correlation to model size, which is another anomalous model behavior our dataset reveals. We discuss the practicality of our metrics in the Appendix.

Next, we explore the potential causes of the color vision effect.

\begin{figure*}[htbp]
    \centerline{
    \begin{subfigure}[b]{0.48\textwidth}  % Adjust width so two figures fit side by side
         \centering
         \includegraphics[width=\textwidth]{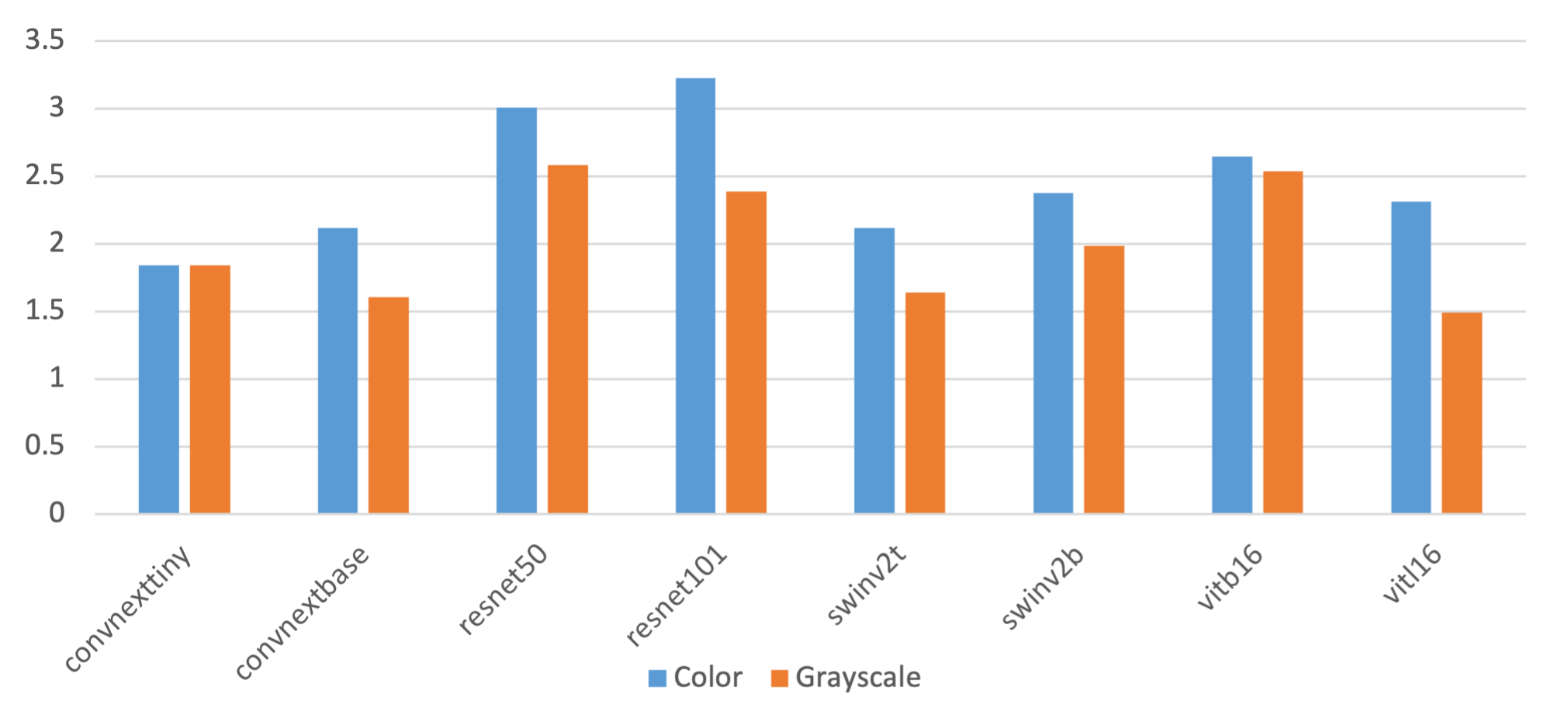}
     \end{subfigure}\hfill
     \begin{subfigure}[b]{0.48\textwidth}  % Adjust width to match the first subfigure
         \centering
         \includegraphics[width=\textwidth]{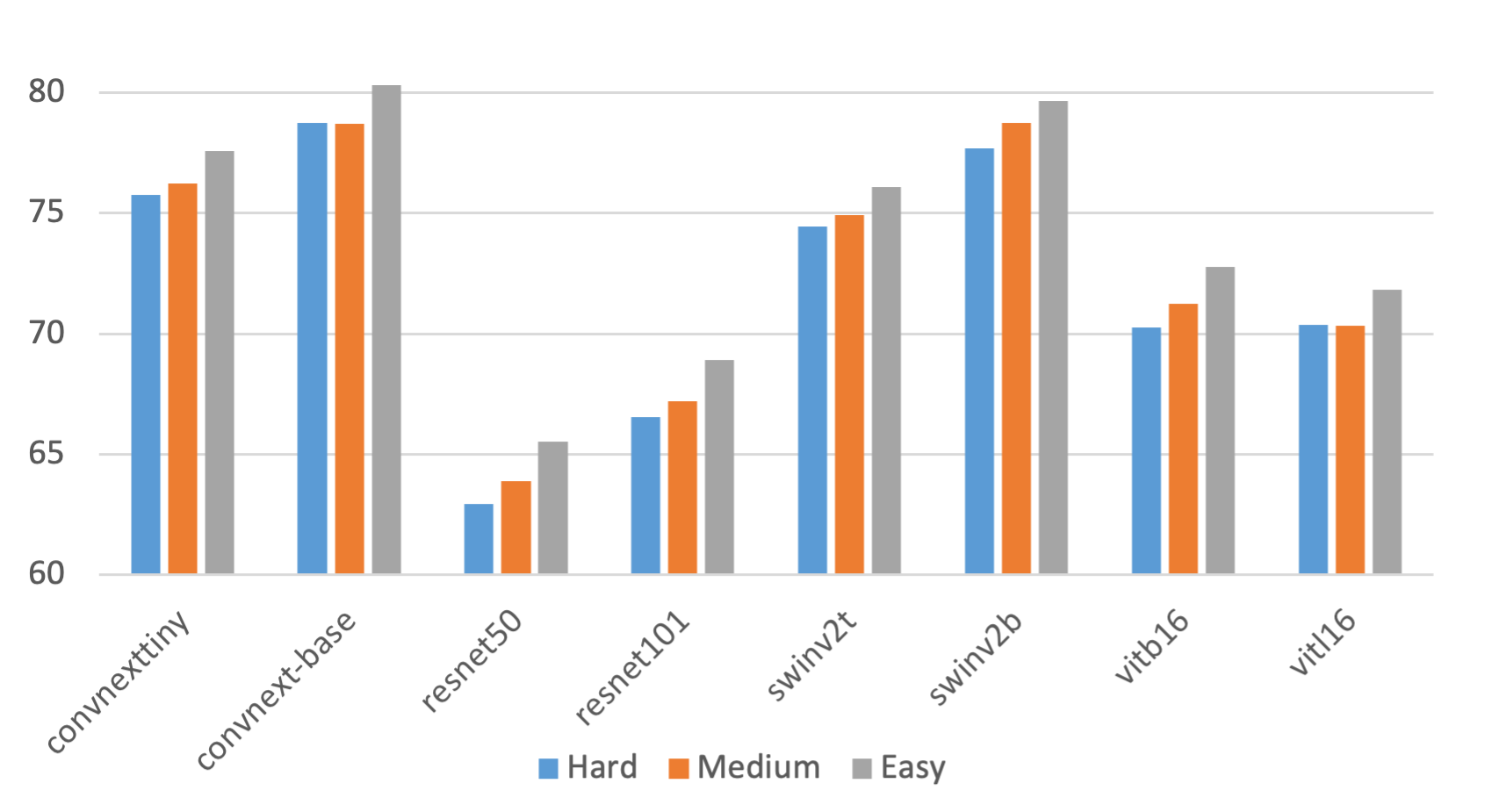}
     \end{subfigure}
     }
    \caption{\textit{Left:} \textbf{Absolute gap (AG) comparisons.} The discrimination effect is more pronounced in color images, supporting our primary finding. \textit{Right:} \textbf{Grayscale ImageNet results (\%).} We still observe slight color vision effects between the CD groups.}
    \label{fig:grayscale_color_grayscale_AG}
\end{figure*}

% \vspace{-4mm}
\paragraph{Less color vision bias on grayscaled images}
To test if chroma is the main cause of the color vision effect, we perform analysis on grayscaled ImageNet. We quantified the absolute gap between the \textsc{Hard} and \textsc{Easy} groups for color and grayscale images, Fig.\ref{fig:grayscale_color_grayscale_AG}-\textit{left}, and observe larger gaps (more pronounced color vision effect) in color images (p-value = $3.9e$-$3$), suggesting a role of chroma in color vision effect. However, we still observe lower performance in harder CD groups (Fig.\ref{fig:grayscale_color_grayscale_AG}-\textit{right}), suggesting the existence of other non-trivial causes. % with models spanning ViTs and CNNs. As in , we still consistently observe a performance decrease within CD groups. However, the color vision effect is more pronounced in colored images, reaffirming our primary finding. %, indicating the reliance of DNN models on color-related features % This is further supported by accuracy percent change analysis (further discussions in the Appendix). Note:: prepare this in appendix and rebuttal
% \vspace{-2mm}
% Grayscale images, although devoid of color information, may retain relative luminance differences. Given our CD group's luminance considerations, the improved grayscale performance may be linked to this factor. Conversely, the larger accuracy drop in grayscale experiments might result from other factors. For instance, grayscale images are out-of-distribution (OOD) for models, potentially explaining the drop compared to colored images; however, this difference should not be directly compared to accuracy drops between CD groups.

% \vspace{-4mm}
\paragraph{Bias persists after controlling luma}
To test if lumination condition lead to the observed color vision effect, we subset images with average luma in certain ranges like $(-0.2, 0.2)$ \& $(-0.4, 0.4)$ and then conduct the same analyses. Table~\ref{tab:luma} shows that when luma is similar, we still observe similar color vision trend (p-val = $5.95e$-$7$ for range (-0.2, 0.2); p-val = $5.95e$-$7$ for range (-0.4, 0.4)), suggesting that chromatic differences play a major role when lumination is controlled.
% \begin{figure}[htbp]
%     \begin{subfigure}[b]{0.9\textwidth}
%          \centering
%          \includegraphics[width=\textwidth]{figures/color_grayscale_abs_gap.png}\vspace{-3mm}
%      \end{subfigure}\hfill
%      \begin{subfigure}[b]{1\textwidth}
%          \centering
%          \includegraphics[width=\textwidth]{figures/grayscale.png}\vspace{-3mm}
%      \end{subfigure}
%     \caption{\textit{Left:} \textbf{Absolute gap (AG) comparisons.} The discrimination effect is more pronounced in color image, supporting our primary finding. \textit{Right:} \textbf{Grayscale ImageNet results (\%).} We still observe slight color vision effect between the CD groups. }
%     \label{fig:grayscale_color_grascale_AG}
%     % \vspace{-2mm}
% \end{figure}

\begin{table}[htbp] 
    \centering
    % \vspace{-4mm}
    \caption{The color vision effect remains after controlling luma. Results reported in Hard/Medium/Easy CD group format.}
    \begin{adjustbox}{width=\linewidth}
    \begin{tabular}{|c|c|c|}
    \hline
    \textbf{Luma range}   & \textbf{(-0.2,0.2)} & \textbf{(-0.4,0.4)} \\\hline
    Resnet50 & 73.59/77.05/77.91 & 73.77/76.51/77.47 \\\hline
    Resnet101 & 74.87/77.60/79.75 & 75.20/77.46/79.09 \\\hline
    ConvnextT & 80.83/82.66/83.17 & 81.14/82.54/83.21\\\hline
    ConvnextB & 82.32/84.20/85.21 & 82.47/83.75/85.06\\\hline
    VitB16 & 79.30/80.89/82.34 & 79.33/80.85/82.20\\\hline
    VitL16 & 78.91/79.65/80.85 & 78.37/79.83/80.93\\\hline
    SwinV2T & 80.09/82.16/82.95 & 80.08/82.19/82.85 \\\hline
    SwinV2B & 82.22/84.31/85.11 & 82.65/84.07/84.97 \\\hline
    \end{tabular}
    \end{adjustbox}
    \label{tab:luma}
    % \vspace{-6mm}
\end{table}

\begin{table}[htbp] 
    \centering
     \caption{High \& low frequency components accuracy (\%). Results reported in \texttt{Resnet50} / \texttt{ViT_L} format.}
    
    \begin{adjustbox}{width=1\linewidth}
    \begin{tabular}{|c|c|c|c|}
    \hline
     % &  \multicolumn{3}{c}{LFC} \\
     \textbf{LFC}  & \textbf{Easy} & \textbf{Medium} & \textbf{Hard} \\\hline
    $r=8$ & 4.04 / 9.64 & 3.44 / 9.12 & 3.23 / 8.04 \\\hline
    $r=16$ & 21.47 / 36.49 & 19.27 / 34.60 & 17.81 / 31.39  \\\hline
    $r=32$ & 51.71 / 64.43 & 49.01 / 62.52 & 47.48 / 60.61 \\\hline
     \textbf{HFC}  & \textbf{Easy} & \textbf{Medium} & \textbf{Hard} \\\hline
    $r=8$ &  44.97 / 51.08 & 44.62 / 51.61 & 45.52 / 52.19\\\hline
    $r=16$ & 13.99 / 9.58 & 14.65 / 10.20 & 15.00 / 11.23 \\\hline
    $r=32$ &  2.01 / 1.82 & 1.85 / 1.64 & 2.57 / 2.67 \\\hline
    \end{tabular}
    \end{adjustbox}
    \label{tab:frequency}
            % \vspace{-5mm}
        % \end{figure}        
    % \vspace{-6mm}
\end{table}

\begin{table}[htbp]

    \caption{Object localization accuracy. The color vision effect exists in another visual recognition task like object localization, not only in classification.}
    
    \begin{center}
    \begin{adjustbox}{width=0.7\columnwidth}
    \begin{tabular}{|c|c|c|c|}
    \hline
    \textbf{CD Group} & \textbf{IoU-30} & \textbf{IoU-50} & \textbf{IoU-70} \\
    \hline
    Easy    & 75.31  & 73.12  & 50.34 \\
    \hline
    Medium  & 74.48  & 72.77  & 50.01 \\
    \hline
    Hard    & 72.94  & 70.80  & 48.97 \\
    \hline
    \end{tabular}
    \end{adjustbox}
    \label{tab:obj_loc}
    \end{center}
\end{table}

% \vspace{-4mm}
\paragraph{Intuition on root cause from frequency analysis} Inspired by \cite{herrmann2022pyramid}, we hypothesize that color vision bias arise from certain frequency components of the images. We conduct frequency analysis and split the frequency components with $r=$ [8, 16, 32]. We split each image into high and low frequency components (HFC/LFC), and test \texttt{Resnet50} and \texttt{ViT_L} on the two components respectively. Results for each CD group are summarized in Tab.~\ref{tab:frequency}. We observe that when presented with only the LFC of images, models consistently perform worse in \textsc{Hard} group, and surprisingly, on HFC both models show better performance for \textsc{Hard} group. Since LFC results resemble what we observe for the original images, we hypothesize that the color vision bias is predominately driven by the LFC in the images. Furthermore, as we already observed lower performance in harder groups on the grayscale experiments (Fig.~\ref{fig:grayscale_color_grayscale_AG}) and considering LFC usually contains most of the color information \cite{gonzalez2009digital}, we conjecture that the color vision bias may be relevant to other low-level features that models pick up, beyond mere colors. \textbf{Consequently, mitigating this bias is non-trivial, as also supported by experiments from previous sections (Tab.~\ref{tab:AG} \& Fig.~\ref{fig:normal_vs_adv} \&  Fig.~\ref{fig:compare_training_data}), which also substantiate the merit of our dataset as this phenomenon cannot be explored before.} 
% \cite{SHANG2024210} also show LFC usually contains most of the color information, but i think it is already common knowledge
% if reviewer asks we can also say fig.7 show changing pre-training data doesnt help either.
% indicating that CV models perform similarly to human vision \ming{do we want this?}.

%% \input{fig_std_imagenet}
% Faye: replace the above .tex with a table + a figure containing two panels as below:

\begin{figure*}[t]
    \centering
    \begin{minipage}[b]{0.48\textwidth}
        \centering
        \includegraphics[width=0.8\textwidth]{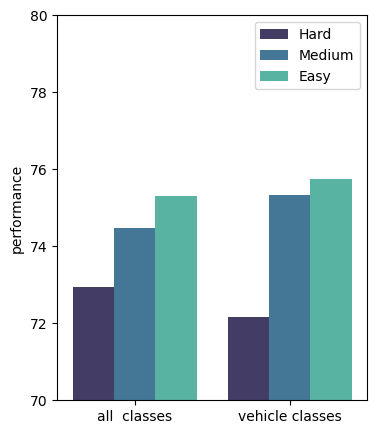}
    \end{minipage}\hfill
    \begin{minipage}[b]{0.48\textwidth}
        \centering
        \includegraphics[width=0.85\textwidth]{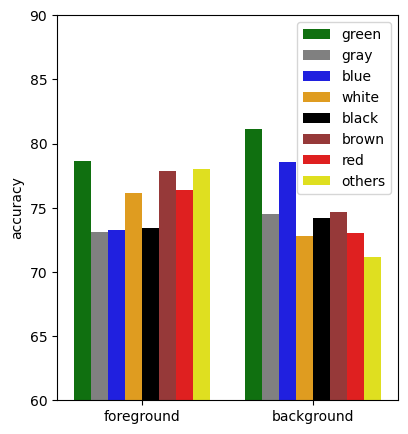}
    \end{minipage}
    \caption{\textit{Left:} Object localization performance on ImageNet (threshold = IoU-30) — vehicle classes are more affected by the color vision effect. \textit{Right:} Background and foreground colors as spurious correlations; green objects/background are favored by \texttt{Resnet50}.}
    \label{fig:std_imagenet}
\end{figure*}

\begin{figure*}
    \centering
    \includegraphics[width=1\linewidth]{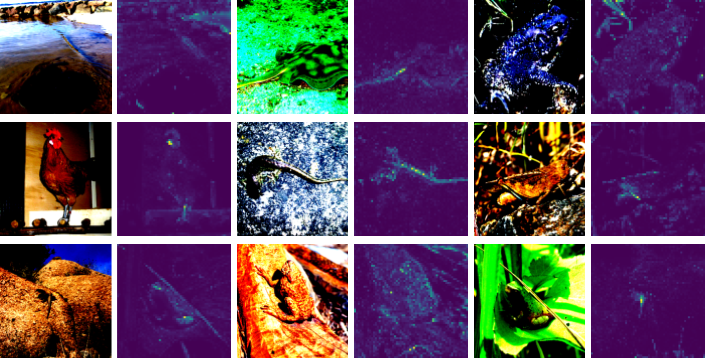}
    \caption{Examples misclassified by models their respective attention maps. Though the objects are easy for humans to recognize, we believe the models' failure stem primarily from the small difference in background and foreground colors.}
    \label{fig:failure}
\end{figure*}

\paragraph{Qualitative results on failure cases} We present nine common failure cases across multiple state-of-the-art models such as ConvNeXt and Swin-Transformer from the \textsc{HARD} group in Fig.~\ref{fig:failure}. These failure cases are deliberately selected to represent scenarios where human visual recognition remains robust, yet the models exhibit significant shortcomings. 

Accompanying each image is the corresponding attention map, though it may not be the best interpretibility technique, we observe that the model may not be able to attend to the right object areas in whole, or sometimes the object is blended in the background. Thus, we believe that the models' misclassifications stem primarily from the small color difference between foreground and background elements, which appear to impede the models' ability to effectively discern and isolate salient features. This phenomenon underscores the ongoing challenges in developing robust computer vision systems capable of matching human perceptual acuity across diverse visual scenarios.

\subsection{Practicality of Proposed sCVR \& sCR Metrics}

We want to highlight the generality of how sCVR was formulated. sCVR is defined to measure the variability of model performance between subgroups while taking accuracy into account. In this work, we have 3 different color discrimination groups per our design; however, in other scenarios, researchers may be interested in a different factor, so they have different subgroups. Therefore, we can design a new metric in a general way, such as sCVR to broadly assess the between-subgroup variability. For instance, we similarly defined an sCR (formula above) for measuring robustness to image corruption (the y-axis in Fig.~\ref{fig:std_imagenet}-\textit{left}). We believe that solely using ``total accurac'' to evaluate models does not provide sufficient granularity, and a performance variability metric is needed for any study design that involves subgrouping. A simple statistical variance does not suffice due to subgroup sample imbalances. We also designed sCVR along with sCR to account for total accuracy since it is commonly regarded as the primary metric measuring model performance.

To validate the practicality of our metric, we want to highlight that in Fig.~\ref{fig:std_imagenet}-\textit{left}, our finding on model robustness to image corruption is consistent with \cite{hendrycks2019robustness} — size-wise, larger models (represented by larger ticks) have higher sCR (more robust). We believe that sCVR and sCR can serve as general metrics for model robustness.

For both sCVR and sCR, we reference the definition of population standard deviation (divided by w, not w-1), rather than sample standard deviation, since our goal is to quantify a variation measurement using all the groups rather than making a specific statistical inference/estimation.

One can also consider model sizes, but that is out of the scope of our work.

\section{Other Utilities: Object Localization and Spurious Correlation}
Our \textsc{ColorSense} has enormous potential for a wide range of applications in computer vision. We already show its usefulness for perception task such as image classification, where color plays a critical role in recognizing objects in an image. We conduct another key visual recognition task, object localization (OL). Other than that, we can study color as a spurious correlation to analyze subgroup performance. In general, \textsc{ColorSense} is a valuable tool for any task that studies color discrimination in computer vision.

\paragraph{Object localization is similar to classification}
We apply the same analysis procedure in Sec.~\ref{sec:results_imagenet} to object localization \cite{bai2022weakly}. We report top-1 localization accuracy with three different IoU thresholds (30, 50, 70). We summarize the results on \textsc{ColorSenese-ImageNet} in Tab.~\ref{tab:obj_loc}. We observe similar performance differences as in classification task, where the model tends to perform better as the CD group becomes easier. And this trend happens across all three different IoU thresholds. This again bolsters our finding of the color vision effect in machine vision.

\paragraph{High-stakes examples in object localization} Similar to classification, we conduct a case study on land vehicle classes for OL due to their resemblance to a self-driving setup. We also observe larger AG between CD groups, meaning an amplified effect of color vision. Fig.~
\ref{fig:std_imagenet}-\textit{left} shows the IoU-30 result and we observe similar trends using other thresholds. We emphasize the worst-case situation in high-stakes tasks. For the \textsc{Hard} group, we observe worse performances on vehicle classes than all classes in all three thresholds, suggesting higher safety risks in those conditions and the necessity of risk mitigation. We believe our findings can be extended to other self-driving studies.

\paragraph{Background and foreground color as spurious correlations}
Another application of our dataset is to leverage the background/foreground labels as spurious contextual correlations. For example, on \texttt{ResNet50}, we observe ``subgroup degradation," uneven performances across subgroups, for both foreground and background. For instance, \texttt{ResNet50} tends to bias in favor of green objects and objects in a green background (Fig.~\ref{fig:std_imagenet}-right). Also, we observe similar trends across all models, showing the universality of this bias. The complete plot for all models is shown in Appendix.

\section{Limitations}
We study color vision in the scope of static visual perception, and it would be interesting to extend the analysis to dynamic visual perception such as video data. 

Furthermore, following our extensive analyses in Sec.~\ref{sec:ablations}, to address the color vision phenomenon that our work revealed in modern DL architectures, only advanced training such as data augmentation shows marginal benefit, and it would be a promising future direction to develop a bespoke data augmentation method to enhance the mitigation of color vision in modern DL architectures.

\section{Discussion}
Our work presents the development and demonstrates the utilities of our curated \textsc{ColorSense} dataset, which is a versatile tool for (1) investigating the impact of color vision on various visual recognition tasks, (2) featuring compatibility with OOD environmental and artificial noise datasets, and (3) controlling for spurious correlations. We argue that current literature on machine vision evaluations does not fully explore the role of color vision in object recognition and show that our framework provides a more comprehensive approach to evaluating machine visual recognition.

% \vspace{-2mm}
Our work demonstrates that current machine vision is susceptible to color vision bias, regardless of DNN architecture, training methods (e.g. self-supervised, contrastive, source data, etc.), size of the model, or controlling factors like luma. We provide quantitative evidence that machine vision is affected by color vision in the form of performance reduction in harder CD groups, and also show that advanced training procedures for robustness enhancement only marginally mitigate the color vision effect. In addition, qualitative examples suggest the background and foreground similarity impedes correct classification when the examples are easy for humans. Moreover, we reason and provide intuition about such a non-trivial phenomenon with frequency analysis, which suggests that more innovations are needed to improve model robustness to color to align with human vision. Further, in the high-stakes use cases studied, the color vision effect may be more obvious. Lastly, CNNs such as \texttt{Convnext} can be as powerful as transformers, as similarly observed in \cite{smith2023convnets}. 

We believe that in the era of foundation models with various scaling laws, our dataset and analysis framework are valuable and can push the boundaries of benchmarking in visual recognition. 

\section{Potential Broader Impact}
% \vspace{-2mm}
The exploration of machine-human alignment stands as a highly coveted focus in research \cite{openai2022alignment}. It is anticipated that advancements in our understanding of human vision \cite{manahilov2003temporal, dapello2020simulating, safarani2021towards} will pave the way for notable breakthroughs in deep learning, specifically addressing the color vision bias in models. Our \textsc{ColorSense} dataset is positioned as a versatile and valuable resource for the comprehensive assessment of newly proposed models and training methods \cite{arjovsky2019invariant, ben2000robust}. As demonstrated in previous sections, this dataset can not only be used to study model biases but also contribute to heightened awareness of safety, spurious signals, and fairness concerns, thereby enriching the field of Machine Learning community.

{
\small
\bibliography{main}
}

% \section{Figure requirements}\label{FAT}
% \paragraph{Positioning Figures and Tables} Place figures and tables at the top and 
% bottom of columns. Avoid placing them in the middle of columns. Large 
% figures and tables may span across both columns. Figure captions should be 
% below the figures; table heads should appear above the tables. Insert 
% figures and tables after they are cited in the text. Use the abbreviation 
% ``Fig.~\ref{fig}'', even at the beginning of a sentence.

% \begin{figure}[htbp]
% \centerline{\includegraphics{fig1.png}}
% \caption{Example of a figure caption.}
% \label{fig}
% \end{figure}

% Figure Labels: Use 8 point Times New Roman for Figure labels. Use words 
% rather than symbols or abbreviations when writing Figure axis labels to 
% avoid confusing the reader. As an example, write the quantity 
% ``Magnetization'', or ``Magnetization, M'', not just ``M''. If including 
% units in the label, present them within parentheses. Do not label axes only 
% with units. In the example, write ``Magnetization (A/m)'' or ``Magnetization 
% \{A[m(1)]\}'', not just ``A/m''. Do not label axes with a ratio of 
% quantities and units. For example, write ``Temperature (K)'', not 
% ``Temperature/K''.

% \newpage
\newpage
\appendix

\subsection{Color Pairs}

We consult experts in the school of fine arts and use the WCAG ratio \cite{w3c2008WCAG} as a guide and threshold to define the color discrimination groups. Tab.~\ref{tab:contrast_group_stats} summarizes the statistics of our grouping and Tab.~\ref{tab:color_pair} lists the color pairs and the corresponding difficulty level.
If either foreground or background color is labeled as `others,' we assign `others' as its color discrimination group as well.  We follow experts' suggestions to use pairs with 1 as \textsc{Hard} (same background and background), pairs with ratios larger than the minimum WCAG ratio, level AA \cite{w3c2008WCAG}, as \textsc{Easy} (note that pairs in \textsc{Easy} are also ranked the top $\frac{1}{3}$ of scores of all pairs), and those in the middle go to \textsc{Medium}.

\begin{table}[t]
    \centering
    % \vspace{-3.5mm}
    \caption{Statistics of \textsc{ColorSense} by color discrimination group.}
    \begin{adjustbox}{width=0.8\linewidth}
    \begin{tabular}{|c|c|c|c|c|}
    \hline
        \textbf{CD Group} & \textbf{\#pairs} & \textbf{ImageNet} & \textbf{CIFAR10} \\\hline
        Hard & 7 & 9121 & 1690 \\\hline
        Medium & 12 & 15835 & 3433 \\\hline
        Easy & 9 & 18503 & 4126\\\hline
        Others & 8 & 6541 & 751 \\\hline
        
    \end{tabular}
    \end{adjustbox}
    \label{tab:contrast_group_stats}
    \vspace{-2mm}
\end{table}

%% keeping cifar100 stats
% CD Group & \#pairs & ImageNet & CIFAR10 & CIFAR100 \\\midrule
%         Hard & 7 & 9121 & 1690 & 1498\\
%         Medium & 12 & 15835 & 3433 & 3699\\
%         Easy & 9 & 18503 & 4126 & 3905\\
%         Others & 8 & 6541 & 751 & 898 \\\bottomrule
% below was old color contrast groups
% \begin{table}[h]
%     \centering
%     \begin{tabular}{ccccc}
%     \toprule
%         \multirow{2}{*}{Contrast level} &  \multirow{2}{*}{\#pairs} & \multicolumn{3}{c}{\#instances} \\
%          & & ImageNet & CIFAR10 & CIFAR100 \\\midrule
%         High & 7 & & 2820 & 2866\\
%         Mid & 6 & & 2124 & 1656\\
%         Low-mid & 9 & & 2615 & 3082\\
%         Low & 7 & & 1690 & 1498\\
%         Others & 8 & & 751 & 898 \\\bottomrule
%     \end{tabular}
%     \caption{Statistics of ImageNet-CoCo \& CIFAR-CoCo by color contrast groups.}
%     \label{tab:contrast_group_stats}
%     % \vspace{-5mm}
% \end{table}

\begin{table}[h]
    \centering
    \begin{tabular}{|c|c|c|c|}
    \hline
        \textbf{Background} & \textbf{Foreground} & \textbf{WAGC ratio} & \textbf{\textsc{ColorSense}} \\
        \hline
green & others & n/a & Others\\\hline
gray & others & n/a & Others\\\hline
blue & others & n/a & Others\\\hline
white & others & n/a & Others\\\hline
black & others & n/a & Others\\\hline
brown & others & n/a & Others\\\hline
red & others & n/a & Others\\\hline
others & others & n/a & Others\\\hline
green & green & 1.0 & Hard\\\hline
gray & gray & 1.0 & Hard\\\hline
blue & blue & 1.0 & Hard\\\hline
white & white & 1.0 & Hard\\\hline
black & black & 1.0 & Hard\\\hline
brown & brown & 1.0 & Hard\\\hline
red & red & 1.0 & Hard\\\hline
green & gray & 1.33 & Medium\\\hline
green & white & 1.37 & Medium\\\hline
blue & brown & 1.39 & Medium\\\hline
brown & red & 1.55 & Medium\\\hline
gray & white & 1.82 & Medium\\\hline
blue & red & 2.15 & Medium\\\hline
gray & red & 2.2 & Medium\\\hline
blue & black & 2.44 & Medium\\\hline
green & red & 2.91 & Medium\\\hline
black & brown & 3.39 & Medium\\\hline
gray & brown & 3.41 & Medium\\\hline
white & red & 4.0 & Medium\\\hline
green & brown & 4.52 & Easy\\\hline
gray & blue & 4.72 & Easy\\\hline
black & red & 5.25 & Easy\\\hline
white & brown & 6.2 & Easy\\\hline
green & blue & 6.26 & Easy\\\hline
blue & white & 8.59 & Easy\\\hline
gray & black & 11.54 & Easy\\\hline
green & black & 15.3 & Easy\\\hline
white & black & 21.0 & Easy\\\hline
        
    \end{tabular}
    \caption{\textbf{Color pairs.}}
    \label{tab:color_pair}
\end{table}

% \section{Background/Foreground as Spurious Correlation Full Plot}
\begin{figure*}[t]
     \begin{subfigure}[t]{\textwidth}
         \centering
         \includegraphics[width=\textwidth]{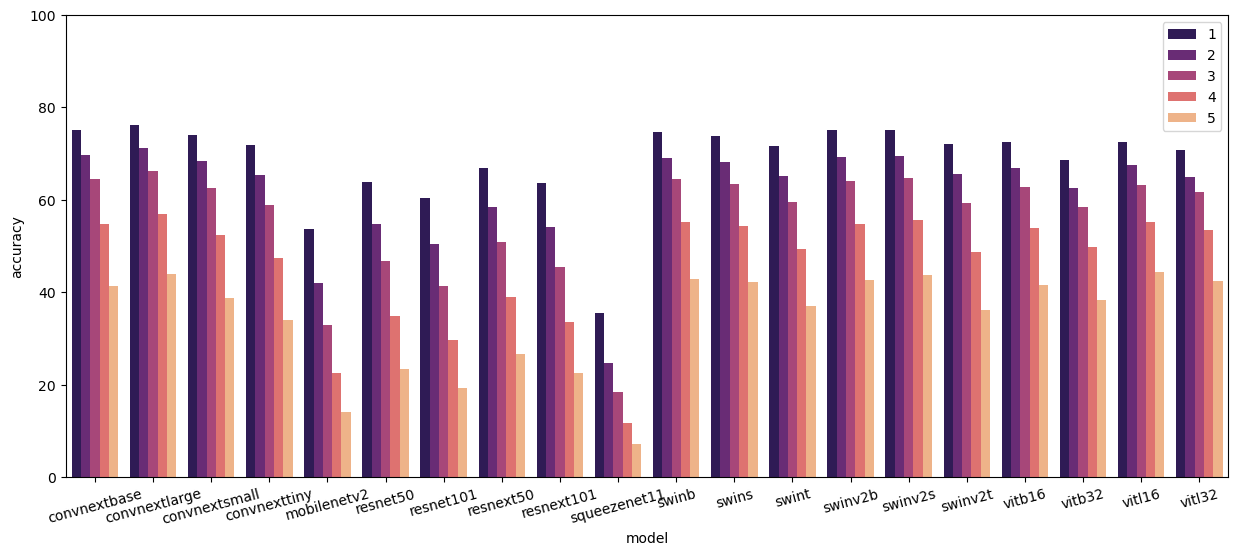}
     \end{subfigure}
     \caption{\textbf{Performance of each architecture on ImageNet-C by severity.} As an another way to validate our \textsc{ColorSense} dataset, the color discrimination groups similarly affect performance as corruption severity (see Fig.~\ref{fig:bar_imagenet} for comparison).}
     \label{fig:supp_allseverity}
     % \vspace{-5mm}
\end{figure*}

\begin{figure*}[t]
     \begin{subfigure}[t]{1\textwidth}
         \centering
         \includegraphics[width=\textwidth]{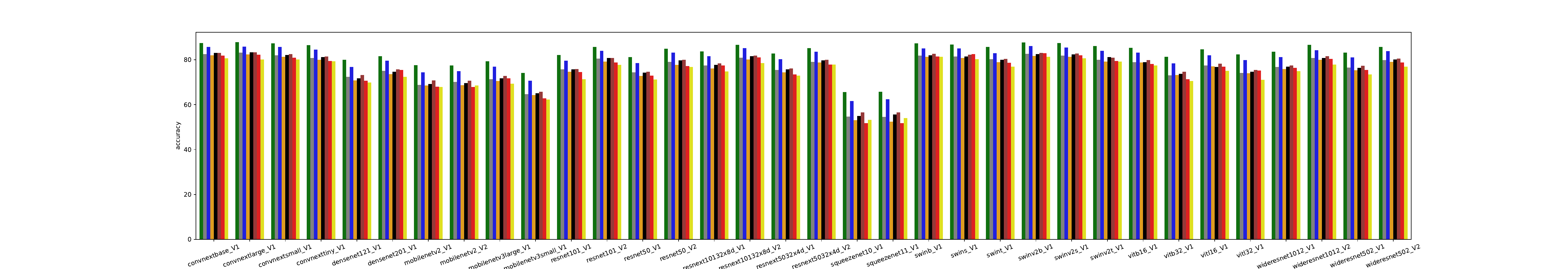}
         \caption{Background color as a spurious correlation.}
     \end{subfigure}
     \begin{subfigure}[t]{\textwidth}
         \centering
         \includegraphics[width=\textwidth]{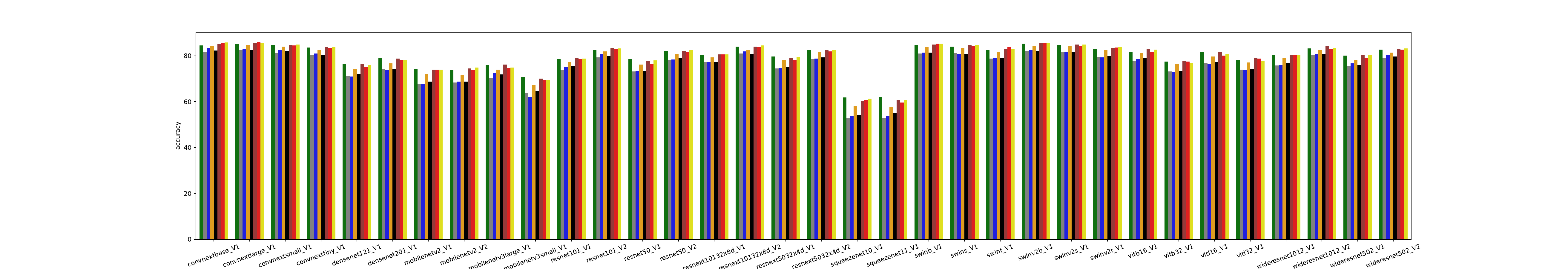}
         \caption{Foreground color as a spurious correlation.}
     \end{subfigure}
     \caption{Background/Foreground color as a spurious correlation. The ``subgroup degradation'' phenomenon appears with all models.}
     \label{fig:supp_background}
     % \vspace{-5mm}
\end{figure*}

\subsection{Vehicle Classes}

The 49 land vehicle classes from ImageNet we considered in this work are: \textit{Model T, ambulance, amphibian, beach wagon, bicycle-built-for-two, bobsled, cab, convertible, crane, dogsled, fire engine, forklift, garbage truck, go-kart, golfcart, grille, half track, harvester, horse cart, jeep, jinrikisha, lawn mower, limousine, minibus, minivan, moped, motor scooter, mountain bike, moving van, oxcart, passenger car, pickup, plow, police van, racer, recreational vehicle, school bus, snowmobile, snowplow, sports car, streetcar, tank, thresher, tow truck, tractor, trailer truck, tricycle, trolleybus, unicycle}.

% \paragraph{A short note on human vision test.}  For humans, to measure vision perceptual ability, a common routine is to perform the \textit{Snellen's test}, where people are asked to discriminate black alphabets in a white background from a certain distance (Fig.~\ref{fig:big_picture} \textit{left-middle}). This test measures our \textit{visual acuity}, or sharpness of vision, the ability to distinguish shapes and the details of objects \cite{visual_acuity_porter, owsley1987contrast}. However, in modern times, ophthalmologists have discovered that a 20/20 vision does not indicate perfect vision \cite{visual_acuity_porter, owsley1987contrast, gregory2015}, and have proposed the \textit{contrast sensitivity test} to measure another aspect of visual function and the purpose is to investigate if people can discriminate finer and finer increments of contrast. Experts deem contrast sensitivity test a more complete assessment of vision \cite{owsley1987contrast}.

\subsection{Small Scale Human Study}

We record 4 participants’ reaction times for a 3-class classification task (n02056570, n02128757, n03095699). We instruct subjects to classify each image into the 3 classes within 2.5 seconds; otherwise, the result of the specific image will be deemed incorrect. 

\begin{figure*}[t]
    \begin{subfigure}[b]{0.5\textwidth}
         \centering
         \includegraphics[width=\textwidth]{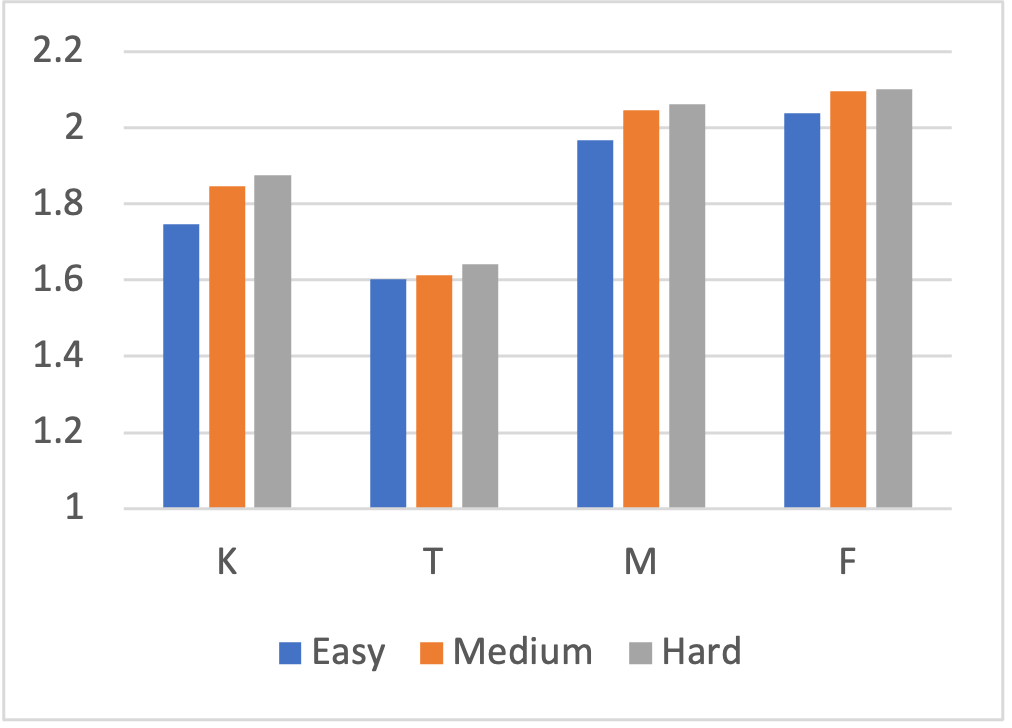}
     \end{subfigure}\hfill
     \begin{subfigure}[b]{0.5\textwidth}
         \centering
         \includegraphics[width=\textwidth]{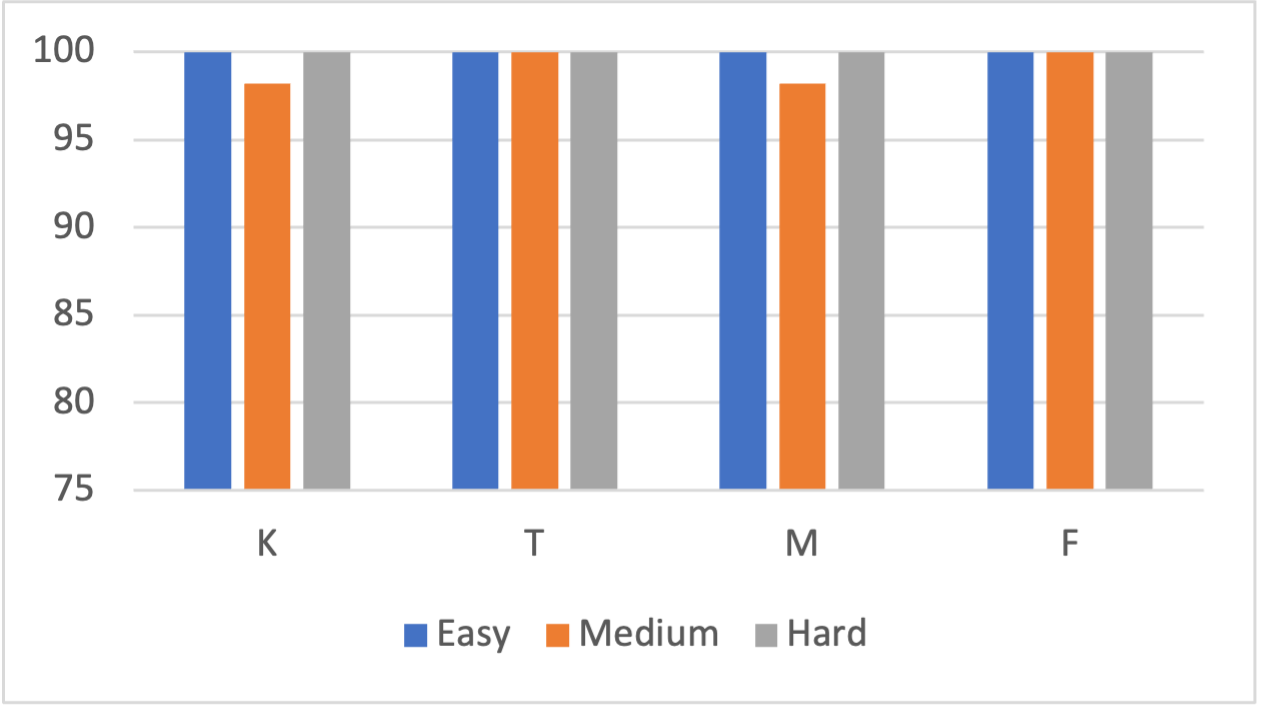}
     \end{subfigure}\vspace{-2mm}
    \caption{\textbf{Human subject experiments.} \textit{Left:} average response time (s) for each color discrimination group on class n02056570, n02128757, n03095699. \textit{Right:} in our experiments, human subjects are not affected by color vision in terms of accuracy when given enough time.
}
    \label{fig:human_study}
    \vspace{-4.5mm}
\end{figure*}

Fig.~\ref{fig:human_study} shows that humans also take a slightly longer time to classify objects for harder CD groups (p-val = $7.716e$-$4$), confirming that color vision plays a role in human vision too, but in the form of response time. We conclude that humans are more robust to color vision than DNNs, as measured by classification accuracy, and indeed, we acknowledge that reaction time can be used as a metric to better understand human’s dependence on color vision. 

Note that this human experiment is meant to verify the conclusion of the literature about humans being robust to color vision groups (Sec.4.1) \cite{dicarlo2012does, zoccolan2007trade, potter1976short, intraub1980presentation, rubin1992reading, logothetis1996visual, thorpe1996speed, edelman1999representation, rousselet2004parallel,Geirhos2020BeyondAQ} and this work's main foci remain on the evaluation of DNN models.

\subsection{Dataset Release}
The \textsc{ColorSense} datasets will be released upon acceptance of this paper.

% \section{DINO results}

\subsection{Label Quality and Procedure}

The color annotations are labeled by an individual with a proficient understanding of computer vision to ensure consistency, and all annotations undergo two rounds of verification. As a quality assurance measure, two individuals with substantial technical expertise assessed 500 randomly selected images. The level of consensus reached was 92.4 percent, and after further discussion, an additional 4.8 percent of images initially lacking agreement aligned with the labels we employed for experiments. The rate of disagreement is lower than the typical error rate found in contemporary datasets \cite{northcutt2021pervasive}. We intend to make the dataset publicly available and encourage the community to contribute updates to the color labels.

 \paragraph{Human annotation} Since the main aim of this work is to understand if machine vision is affected by color similarly as human vision is, we decide that it makes more sense to use human labeling as our ground truth, rather than a pixel value based metric. In addition, pixel-based color contrast calculation can be strongly biased by pixel selection. Manually selected pixels can be located in various places on the object, and confounding factors such as lumination condition can strongly bias the foreground-background color contrast quantification. Finally, since the labeller is only asked to choose between 8 distinct groups and human vision is robust to color vision and would not be strongly affected by lumination condition, the chances that humans mislabeled a non-black object as black is low when compared with that when using pixel values. To summarize, we think human labeling is more suitable and way more robust for the purpose of our study.

\paragraph{Guideline} We emulate similar labeling criteria to \cite{chiu2022better}, Sec. 3.2.

\begin{figure*}[t]
     \begin{subfigure}[t]{0.49\textwidth}
         \centering
         \includegraphics[width=\textwidth]{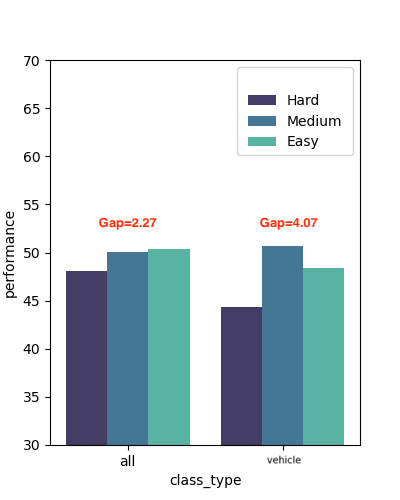}
         \caption{IoU-70}
     \end{subfigure}\hspace{0.1em}
     \begin{subfigure}[t]{0.49\textwidth}
         \centering
         \includegraphics[width=\textwidth]{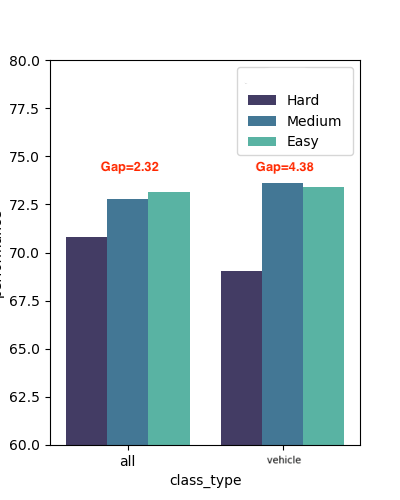}
         \caption{IoU-50}
     \end{subfigure}
     \caption{\textbf{Object localization accuracy (\%) on all ImageNet classes vs. land vehicle classes.} The performance gap between \textsc{Hard} and \textsc{Easy} groups for vehicle classes are consistently larger than all classes, suggesting vehicle classes are more affected by the color vision effect and therefore possessing safety-related issues. The results are also consistent with the case study in classification task (Sec.~\ref{sec:ablations}).}
     \label{fig:supp_objloc_5070}
     % \vspace{-5mm}
\end{figure*}

% \vspace{-3mm}
\subsection{Results on \textsc{ColorSense-CIFAR10}}
\begin{figure}[t]
     % \begin{subfigure}[b]{0.52\textwidth}
         \centering
         \includegraphics[width=0.5\textwidth]{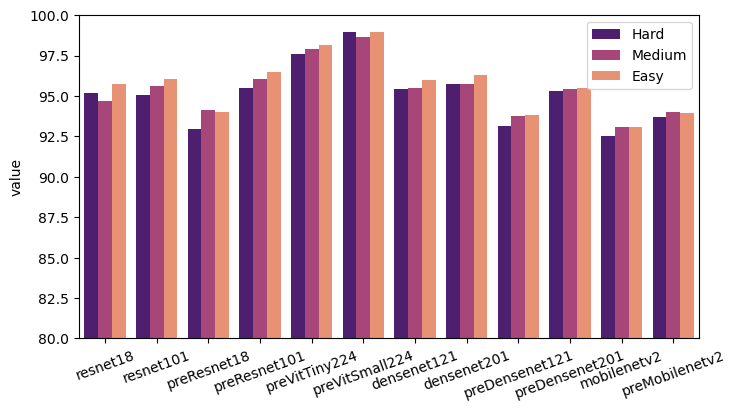}
     % \end{subfigure}\vspace{-4mm}
    \caption{CIFAR10 performances.}
    \label{fig:cifar10_bar}
    % \vspace{-4.5mm}
\end{figure}

% \vspace{-2mm}
We apply Resnet, Densenet \citep{huang2017densely} and ViT as backbones to evaluate on \textsc{ColorSense-CIFAR10} (CIFAR10-B is from \cite{chiu2022better}). The training details of these models are provided in the Appendix. In general, we observe a similar trend of having better performance in easier groups as on ImageNet. Fig.~\ref{fig:cifar10_bar} summarizes the performances of both pre-trained models (pre-\{model\}) and those trained from scratch. Interestingly, not all models are affected consistently by color vision, and pre-training on ImageNet also does not help reduce AG consistently. Though the results on CIFAR10 and ImageNet are slightly different, we believe the ImageNet results and findings are more representative of real-world scenarios.

\subsection{Training Details of CIFAR10 Models}
We train Resnets/Densenets/MobileNetV2 for 250 epochs with stochastic gradient descent (SGD), weight decay of 0.0005, an initial learning rate of 0.1 and milestone learning rate decay at epochs [150, 200] with a decay factor of 0.1. For ViTs, we follow the procedure in \cite{dosovitskiy2020image} to fine-tune ViTs (preVit) on resized CIFAR10 with SGD, learning rate of 0.003 and no weight decay. We use these training setups and models to evaluate them throughout the paper. Furthermore, as ViT are pre-trained on Imagenet, for fairness, we also include pre-trained Resnets (preResnet) and Densenets (preDensenet) that emulate the protocol in \cite{BiT} to fine-tune with milestone learning rate decay schedule at one-third and two-thirds of the process. All models are trained with a batch size of 128\footnote{In \cite{BiT} they use a batch size of 512 but we find that the results are not good, and for a fair comparison, we finally choose a batch size as 128.}. In this work, generalization performances are not our foci, so we do not tune any hyperparameters. 

We emphasize that we do not compete for test set accuracy in this work, so generalization performance is not our concern as long as it is reasonable. Here we detail our training procedures and their respective performances.

\paragraph{Vision Transformers (preVit224)} We use \cite{rw2019timm} for ViT implementation. \cite{dosovitskiy2020image} fine-tunes models on CIFAR images resized to resolution 384 and a batch size of 512 for 10k steps (about 25 epochs). Due to our computation-resource limit, we instead resize to resolution 224 and use a batch size of 128. Essentially we prolong the training. Using our recipe, for Tiny ViT, we achieve 97.83\% on CIFAR10, respectively, which is not far away from ViT-B/16 reported in \cite{dosovitskiy2020image}; for Small ViT, we even achieve accuracy 98.73\% on CIFAR10, even better than what is reported in \cite{dosovitskiy2020image}. Note that the model size of Tiny ViT or Small ViT is smaller or equal to what is used in \cite{dosovitskiy2020image}.

\paragraph{Pre-trained Resnets (preResnet)} We use the pre-trained weights released by PyTorch\cite{NEURIPS2019_9015}. The original recipe in \cite{BiT} is to fine-tune Resnets with a batch size of 512 for 10k steps and learning rate decay at $\frac{1}{3}$ and $\frac{2}{3}$ through the process. To have fair procedures as ViTs, we follow the same setups but not resizing. We use a batch size of 128 to fine-tune on CIFAR for 25 epochs and decay learning rate at 8th and 16th epoch. We found using our recipe is much better than using the original recipe: with the original recipe we achieve accuracy 89.49\% on CIFAR10, but our recipe can reach 93.7\%. 
\paragraph{Pre-trained Densenets/MobilenetV2 (preDensenet/preMobilenetV2) } For fair comparison, we follow the procedure as in preResnet and achieve 95.32\%, 93.73\% respectively.
\paragraph{Resnets/Densenets/MobilenetV2} We achieve accuracy 95.16\%/95.85\%/92.73\% on CIFAR10.

\subsection{Compute}
We have a humble compute resource: all experiments are done on 1 NVIDIA V100 GPU.

\end{document}